\documentclass[lettersize,journal]{IEEEtran}
\usepackage{amssymb}
\usepackage{amsmath,amsfonts}
\usepackage{array}
\usepackage[caption=false,font=normalsize,labelfont=sf,textfont=sf]{subfig}
\usepackage{textcomp}
\usepackage{stfloats}
\usepackage{url}
\usepackage{verbatim}
\usepackage{graphicx}
\usepackage{cite}
\usepackage{booktabs}
\usepackage{multirow}
\usepackage{makecell}

\usepackage[table]{xcolor}
\usepackage[hidelinks]{hyperref}

\newcommand{\ie}{\textit{i.e.}}

\begin{document}

\title{T2VAttack: Adversarial Attack on Text-to-Video Diffusion Models}

\author{
Changzhen Li,~\IEEEmembership{Student Member,~IEEE,}
Yuecong Min,~\IEEEmembership{Member,~IEEE,}
Jie Zhang,~\IEEEmembership{Member,~IEEE,}

Zheng Yuan,~\IEEEmembership{Student Member,~IEEE,}
Shiguang Shan,~\IEEEmembership{Fellow,~IEEE,}
Xilin Chen,~\IEEEmembership{Fellow,~IEEE,}

\thanks{
Changzhen Li and Shiguang Shan are with Hangzhou Institute for Advanced Study, UCAS, school of Intelligent Science and Technology, China, and also with the State Key Laboratory of AI Safety, Institute of Computing Technology (ICT), Chinese Academy of Sciences (CAS), Beijing 100190, China (e-mail: changzhen.li@vipl.ict.ac.cn; sgshan@ict.ac.cn).

Yuecong Min, Jie Zhang, Zheng Yuan, and Xilin Chen are with the State Key Laboratory of AI Safety, Institute of Computing Technology (ICT), Chinese Academy of Sciences (CAS), Beijing 100190, China, and also with the University of Chinese Academy of Sciences (UCAS), Beijing 100049, China (e-mail: minyuecong@ict.ac.cn; zhangjie@ict.ac.cn; zheng.yuan@vipl.ict.ac.cn; xlchen@ict.ac.cn).}
}

\markboth{Journal of \LaTeX\ Class Files,~Vol.~14, No.~8, August~2021}%
{Shell \MakeLowercase{\textit{et al.}}: A Sample Article Using IEEEtran.cls for IEEE Journals}

\IEEEpubid{0000--0000/00\$00.00~\copyright~2021 IEEE}

\maketitle

\begin{abstract}

The rapid evolution of Text-to-Video (T2V) diffusion models has driven remarkable progress in generating high-quality, temporally coherent videos from natural language descriptions. Despite these achievements, the robustness of T2V models against subtle prompt perturbations remains largely unexplored. 
In this paper, we present T2VAttack, a systematic framework for evaluating the adversarial robustness of T2V diffusion models, targeting semantic alignment and temporal dynamics as the principal dimensions of video generation quality.
We formalize the attack as a unified optimization problem that maximizes the degradation of video evaluation scores under imperceptibility constraints, and instantiate it with two video-specific objectives: a semantic objective that measures video-text alignment in a joint embedding space, and a temporal objective that characterizes inter-frame motion patterns via optical flow.
To solve this problem in the black-box setting, we develop two attack methods: (i) T2VAttack-S, which identifies semantically or temporally critical words in prompts and replaces them with synonyms via greedy search, and (ii) T2VAttack-I, which iteratively inserts optimized words with minimal perturbation to the prompt. 
We further construct T2VAttackBench, a curated prompt benchmark for reliable robustness assessment. 
Extensive experiments on four state-of-the-art T2V models, including ModelScope, CogVideoX, Open-Sora, and HunyuanVideo, reveal that even single-word substitutions or insertions can substantially degrade semantic fidelity and temporal dynamics. These findings expose intrinsic vulnerabilities in the cross-modal alignment and temporal modeling mechanisms of current T2V diffusion models.

\end{abstract}

\begin{IEEEkeywords}
Adversarial attack, Diffusion models, Text-to-video generation, Temporal dynamics.
\end{IEEEkeywords}

\section{Introduction}
\label{sec:intro}

The emergence of Text-to-Video (T2V) diffusion models, capable of generating coherent and high-quality videos from textual prompts, has attracted unprecedented attention over the past two years~\cite{videoworldsimulators2024,wang2023modelscope,esser2023structure,wang2025lavie,chen2024videocrafter2,yang2024cogvideox,long2024videodrafter,wang2025wan}. 
Pioneering works such as Sora~\cite{videoworldsimulators2024}, CogVideoX~\cite{yang2024cogvideox} and Wan~\cite{wang2025wan} have demonstrated remarkable success in simulating complex worlds through minute-long video generation, exhibiting impressive spatiotemporal modeling capabilities. These advances hold transformative potential across various domains, including filmmaking, embodied intelligence, and physical world simulation. 
However, the rapid growth in generative capability has outpaced our understanding of model reliability.
While benign users expect faithful generation, subtly perturbed prompts can severely corrupt the generated content.
A systematic study of T2V adversarial robustness therefore serves several practical purposes. First, it provides an effective red-teaming tool to uncover potential vulnerabilities prior to deployment~\cite{ganguli2022red}, which is increasingly important as prompts routinely pass through third-party channels and are thus exposed to manipulation. Second, during deployment, it establishes a rigorous robustness evaluation protocol that delivers trustworthy security reports for high-risk scenarios~\cite{carlini2017towards,guo2023comprehensive}, complementing standard benchmarks that measure only average-case quality. Third, in the post-deployment phase, it empowers proactive defense paradigms, such as adversarial training~\cite{madry2018towards}, by providing actionable robustness recommendations driven by adversarial examples. Ultimately, these integrated efforts safeguard the generation of safe and trustworthy artificial intelligence content.

\begin{figure*}[!t]
    \centering
    \includegraphics[width=0.98\linewidth]{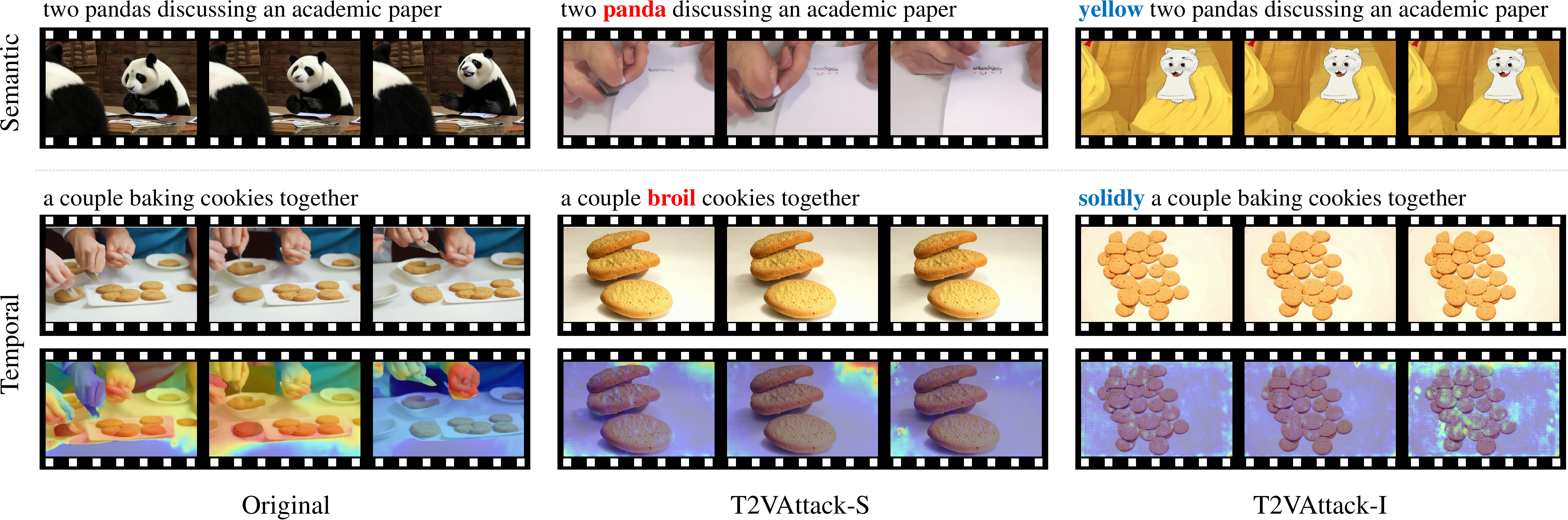}
    \caption{Examples of adversarial attacks on CogVideoX under semantic and temporal objectives.
    The first and second rows show the attack results targeting the semantic and temporal objective, respectively; the third row visualizes the corresponding optical flow magnitudes for the temporal case. Columns show videos generated from the original prompt (left), after T2VAttack-S (middle), and after T2VAttack-I (right).
    The examples highlight that even single-word modifications can cause significant degradation in semantic fidelity and temporal dynamics.
    }
    \label{fig:intro}
\end{figure*}

In contrast, the adversarial robustness of Text-to-Image (T2I) diffusion models has been extensively studied~\cite{liu2023riatig,zhuang2023pilot,yang2024mma}. For example, RIATIG~\cite{liu2023riatig} leverages genetic algorithms \cite{holland1992genetic} to craft stealthy adversarial prompts via word-level perturbations, achieving imperceptible yet effective attacks.
Similarly, QueryFreeAdv~\cite{zhuang2023pilot} demonstrates that character-level perturbations can effectively attack Stable Diffusion models through Projected Gradient Descent (PGD)~\cite{madry2018towards} and greedy search~\cite{temlyakov2011greedy}. 
These approaches pose tangible threats, especially when integrated into tools such as autocomplete plugins, which can generate seemingly natural adversarial prompts that bypass content moderation and yield harmful outputs. 
Nevertheless, such attacks do not transfer readily to video generation. T2I attacks adapted to T2V models induce only marginal degradation, \textit{e.g.}, QueryFreeAdv~\cite{zhuang2023pilot} optimizes 
surrogate objectives in the text-encoder embedding space that prove statistically decoupled from the video-level spatiotemporal representation. 
In essence, T2V generation poses two unique robustness challenges: (1) semantic vulnerability, stemming from multi-frame conditional generation that requires precise cross-modal alignment~\cite{zhang2026trajectory}, and (2) temporal vulnerability, rooted in the complexity of modeling realistic motion dynamics~\cite{wang2025freezevla,chowdhury2025vid}.
To bridge these gaps, we conduct a systematic study of the adversarial robustness of T2V models from both semantic and temporal perspectives, with representative examples illustrated in Fig.~\ref{fig:intro}.

\IEEEpubidadjcol

Our attack objectives are grounded in semantic alignment and temporal dynamics, the two principal evaluation dimensions of video generation~\cite{huang2024vbench}.
For the semantic dimension, prior attacks on T2I models typically employ image-text similarity to measure the semantic deviation of generated images~\cite{zhuang2023pilot}. However, videos convey richer spatiotemporal semantics, such as motion speed and trajectory, that frame-level measures cannot capture. Therefore, we introduce a \textbf{Semantic Attack Objective} that quantifies the alignment between a generated video and its corresponding prompt via video-text similarity.
For the temporal dimension, video generation relies heavily on temporal cues to maintain natural and coherent motion. We thus introduce a \textbf{Temporal Attack Objective} that leverages motion estimation techniques, \ie, optical flow, to characterize motion vectors across neighboring frames.
An effective temporal attack suppresses the motion dynamics specified by the prompt, with a completely static video as the extreme case~\cite{wang2025freezevla,chowdhury2025vid}, thereby degrading the realism of the generated video.

Building on these objectives, we propose \textbf{T2VAttack}, a comprehensive framework for investigating the adversarial robustness of T2V diffusion models. After analyzing four fundamental text editing operations, \ie, substitution, reordering, insertion, and deletion (Table~\ref{tab:attackers_position}), we observe that deletion and reordering operations yield limited attack effectiveness. Consequently, we focus on two more effective word-level adversarial strategies: the substitution attack \textbf{(T2VAttack-S)} and the insertion attack \textbf{(T2VAttack-I)}.
Inspired by textual adversarial methods in Natural Language Processing (NLP) such as TextFooler~\cite{jin2020bert}, T2VAttack-S first identifies the most influential words in a prompt by measuring their importance to the attack objectives, and then replaces them with carefully chosen synonyms via a greedy search to maximize semantic or temporal discrepancies. In contrast, T2VAttack-I iteratively inserts optimized words into the prompt, subtly distorting its semantics or temporal dynamics.
As quantified by our proposed objectives, both strategies significantly degrade semantic fidelity and temporal dynamics while maintaining high semantic and formal similarity to the original prompt. 

To facilitate rigorous evaluation, we construct \textbf{T2VAttackBench}, a high-quality textual dataset describing visually rich and highly dynamic content, built upon manually curated prompts from VBench~\cite{huang2024vbench} and GPT-4o-based prompt rewriting tailored to T2V generation.
We then conduct extensive experiments on four representative T2V models, ModelScope \cite{wang2023modelscope}, CogVideoX \cite{yang2024cogvideox}, Open-Sora \cite{zheng2024open}, and HunyuanVideo \cite{kong2024hunyuanvideo}.
Experimental results demonstrate that our attacks achieve strong effectiveness within a minimal query budget, requiring only a few dozen queries and single-word modifications. These findings reveal critical vulnerabilities of current T2V models, as even imperceptible textual perturbations can severely disrupt their outputs, underscoring the necessity of robustness assessment before real-world deployment.

In summary, our main contributions are as follows:
\begin{itemize}
    \item We formally define the adversarial attack of T2V diffusion models as a unified optimization problem from both semantic and temporal dimensions, highlighting the fundamental differences between attacking video and image generation.
    \item We propose T2VAttack, a general framework for the adversarial evaluation of video generation, which integrates two video-specific attack objectives, two black-box attack methods, a structured evaluation protocol, and a high-quality benchmark dataset.
    \item We conduct extensive experiments on four state-of-the-art T2V models, revealing that even imperceptible single-word substitutions or insertions can severely degrade both semantic fidelity and temporal dynamics, thereby exposing their intrinsic spatiotemporal vulnerabilities.
\end{itemize}

\section{Related Work}
\label{sec:related}

\subsection{Text-to-Video Diffusion Models}
Driven by advancements in Transformers \cite{vaswani2017attention}, Diffusion Models \cite{ho2020denoising}, and Diffusion Transformers \cite{peebles2023scalable}, text-to-video generation has seen remarkable progress \cite{yang2024cogvideox,wang2023modelscope,videoworldsimulators2024,esser2023structure,wang2025lavie,chen2024videocrafter2,long2024videodrafter,zheng2024open,kong2024hunyuanvideo}. T2V diffusion models mainly consist of three key components: a text encoder, a diffusion network, and a video decoder. The text encoder, leveraging pre-trained models such as CLIP~\cite{radford2021learning}, T5~\cite{t5}, and Multimodal Large Language Models (MLLMs)~\cite{liu2023visual}, transforms input prompts into textual embeddings that guide the video generation process. The diffusion network, often implemented with U-Net~\cite{ho2020denoising} or more recent DiT~\cite{peebles2023scalable} series architectures, is the core of the generative process. It samples initial video latent representations from a Gaussian distribution and iteratively denoises the video latents into coherent video representations over multiple timesteps. Finally, the video decoder, typically a Variational Autoencoder (VAE)~\cite{kingma2014auto} or variants like VQ-VAE~\cite{van2017neural} and VQGAN~\cite{esser2021taming}, translates the denoised latent representations back into raw video data.

To generate high-quality videos, various improvements have been proposed. Early works like Make-A-Video~\cite{singer2022make} and Imagen-Video~\cite{ho2022imagen} demonstrate the potential of cascaded diffusion models for video synthesis. To enable multi-view synthesis, Stable Video Diffusion~\cite{blattmann2023stable} extends pretrained text-to-image models with temporal layers and staged training for video generation. More recently, ModelScope \cite{wang2023modelscope}, building upon Stable Diffusion, introduces spatiotemporal blocks to enhance frame consistency and motion smoothness. CogVideoX \cite{yang2024cogvideox} utilizes a 3D VAE for spatial and temporal video compression and employs an expert transformer with adaptive LayerNorm to facilitate deep cross-modal fusion. Open-Sora \cite{zheng2024open} adopts a Spatial-Temporal Diffusion Transformer (STDiT) and a compressive 3D autoencoder to enhance efficiency and flexibility. 
Most notably, HunyuanVideo \cite{kong2024hunyuanvideo}, a significant advancement with 13 billion parameters,
achieves high-quality video synthesis through a combination of meticulous data curation, advanced architecture, and progressive model scaling. Concurrently, commercial models such as Sora~\cite{videoworldsimulators2024}, Pika~\cite{pika2024pika}, Kling~\cite{kuaishou2024kling}, and Gen-3~\cite{runway2024gen3} have showcased impressive video quality with fine-grained control.
In this paper, we select four state-of-the-art models, ModelScope \cite{wang2023modelscope}, CogVideoX \cite{yang2024cogvideox}, Open-Sora \cite{zheng2024open}, and HunyuanVideo \cite{kong2024hunyuanvideo}, as our victim models for adversarial evaluation. Detailed configurations of these models are provided in Appendix~A.

\subsection{Adversarial Attack on Diffusion Models}
Adversarial attacks in computer vision typically involve introducing slight perturbations to input data, causing models to yield incorrect outputs. These attacks are commonly categorized by the attacker's knowledge (white-box attacks \cite{goodfellow2015explaining,madry2018towards,carlini2017towards} vs. black-box attacks \cite{ilyas2018black,liu2024boosting,xu2023best,bai2023query}) and by the target objective (targeted \cite{goodfellow2015explaining,guo2024efficient} vs. untargeted attacks \cite{dong2018boosting}). In NLP, adversarial attacks are further classified by the granularity of perturbations (character-level \cite{gao2018black,eger2019text,yao2024embedding}, word-level \cite{jin2020bert,alzantot2018generating,li2020bert,papernot2016crafting,ren2019generating,hu2024fasttextdodger}, and sentence-level \cite{zhao2018generating,iyyer2018adversarial}). Beyond serving as attack targets, diffusion models have also been leveraged to synthesize visually natural and highly transferable adversarial examples~\cite{liu2025diffprotect,liu2026improving}. Our work mainly focuses on word-level black-box attacks.

Specifically, adversarial attacks on diffusion models can be further categorized based on the location where perturbations are applied~\cite{truong2025attacks}. These include input image-based \cite{salman2023raising,shan2023glaze,zhang2023robustness,yu2024step}, text prompt-based \cite{zhuang2023pilot,zhang2024revealing,liu2023discovering,kou2023character,liu2023riatig,yang2024mma,yang2024sneakyprompt,gao2023evaluating}, and fine-tuning image-based attacks \cite{liang2023adversarial,liang2023mist,zhu2024watermark,van2023anti,kang2025sita}. 
Our work is most closely related to the text prompt-based attacks, which manipulate textual prompts to induce unintended or harmful content. 

Text prompt-based attacks on T2I models have been extensively studied. 
A major goal of these attacks is to misalign the generated image with the original prompt, often by degrading visual quality or altering content semantics. For instance, QueryFreeAdv~\cite{zhuang2023pilot} appends optimized characters to reduce embedding similarity with the original prompt, shifting the generated content without querying the whole model. SAGE~\cite{liu2023discovering} leverages an LLM to generate semantically natural suffixes to mislead the model generation toward unintended content. Other studies, such as RealWorldAdv~\cite{gao2023evaluating}, exploit common typographic and glyph errors as perturbations to explore real-world robustness vulnerabilities, while CharGrad~\cite{kou2023character} applies visually similar character substitutions to discreetly alter outputs.
Another line of text prompt-based attacks aims to bypass safety mechanisms and generate targeted or Not-Safe-For-Work (NSFW) content. TargetedAdv~\cite{zhang2024revealing} performs gradient-based embedding optimization to bypass keyword detectors and generate images from a specific target category. RIATIG~\cite{liu2023riatig} employs genetic algorithms to find imperceptible adversarial prompts that are semantically different yet produce visually similar outputs. Meanwhile, MMA-Diffusion~\cite{yang2024mma} uses a multi-modal objective to bypass both text and image filters, while SneakyPrompt~\cite{yang2024sneakyprompt} applies reinforcement learning to efficiently discover adversarial tokens.
Although these studies collectively reveal that T2I models are highly susceptible to prompt-based adversarial attacks, the adversarial robustness of T2V models remains largely unexplored.
Furthermore, the direct application of these T2I attack methodologies to the video generation domain typically yields a pronounced degradation in attack efficacy. For example, commonly used encoder-level objectives~\cite{zhuang2023pilot} transfer poorly to the video generation setting, where the generation outcome is decoupled from embedding-space displacement.

To the best of our knowledge, several contemporaneous works have investigated jailbreak attacks on T2V models~\cite{liu2025t2v,lee2025jailbreaking}.
For instance, T2V-OptJail~\cite{liu2025t2v} introduces a discrete prompt optimization framework to systematically bypass safety filters for NSFW content generation, whereas SceneSplit~\cite{lee2025jailbreaking} executes black-box jailbreaks by fragmenting harmful narratives into a sequence of superficially benign scenes.
In contrast, our work focuses on evaluating the intrinsic adversarial robustness of T2V models by targeting video-specific vulnerabilities, rather than merely bypassing external safety mechanisms.
Specifically, our attack targets the temporal properties of the video itself, leveraging optical flow for the temporal objective and video-text similarity for the semantic objective, as opposed to the image-text similarity metrics commonly employed in existing jailbreaks~\cite{liu2025t2v}.
Furthermore, our optimization seeks to maximize semantic or temporal discrepancies under general conditions, without being limited to NSFW scenarios, thereby enabling a broader and more systematic assessment of the intrinsic robustness of T2V models.

\clearpage

\begin{figure*}[!t]
    \centering
    \includegraphics[width=0.98\linewidth]{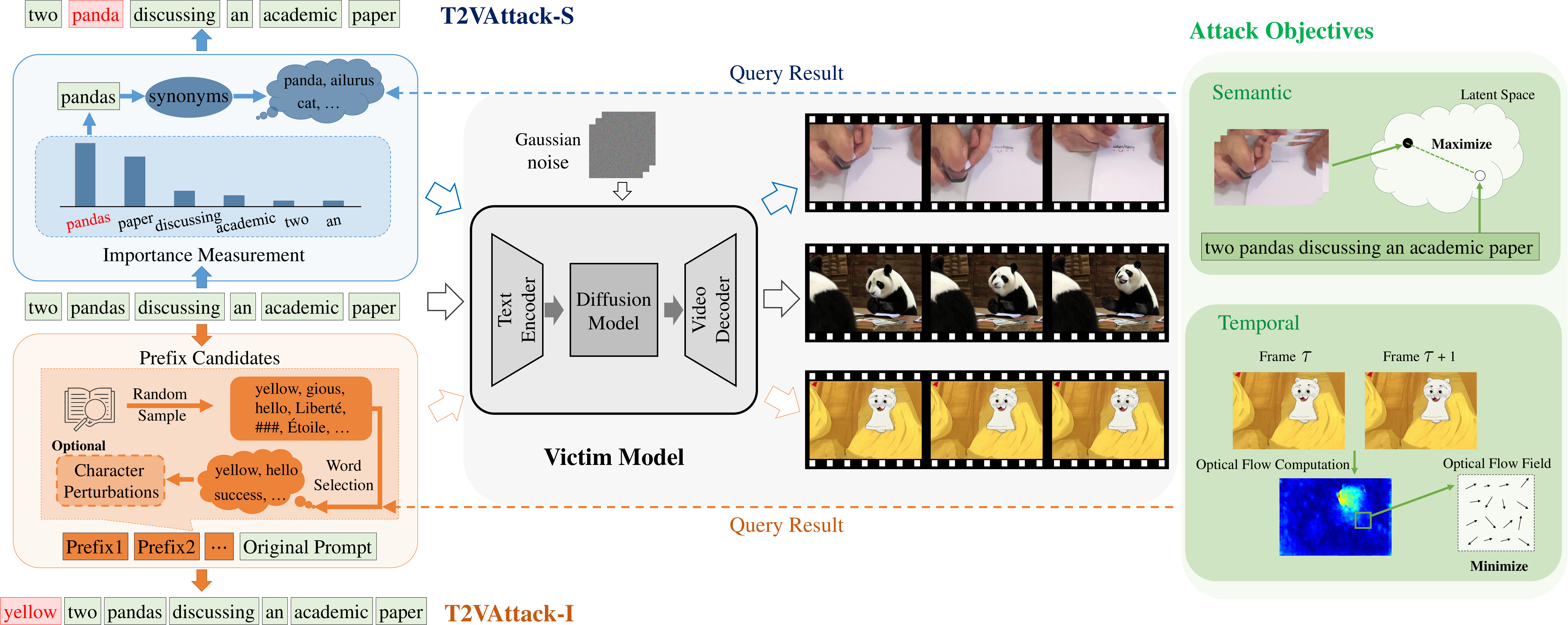}
    \caption{Overview of the adversarial attack pipeline for T2V models. This pipeline illustrates the key process of two attack methods (T2VAttack-S and T2VAttack-I), two attack objectives, and the targeted T2V victim models. It systematically demonstrates how minor prompt perturbations can disrupt the fidelity and dynamics in the generated videos.
    }
    \label{fig:pipe}
\end{figure*}

\section{Method}
\label{sec:method}
\subsection{Problem Formulation}
Let $G$ denote a text-to-video model that takes a textual prompt $X$ as input and produces a video $G(X)$ expected to be semantically aligned with $X$.
The attacker aims to construct an adversarial prompt $X'$ that maximally degrades the quality of the generated video $G(X')$, while ensuring that the perturbation is imperceptible, \ie, $X'$ remains both semantically and formally similar to $X$.
Formally, this can be formulated as the following constrained optimization problem:
\begin{equation}
\max_{X'} \ \mathcal{L}\Big(f(G(X)), f(G(X'))\Big) \quad \text{s.t.} \quad g(X, X') \geq \epsilon, 
\label{equ:formulation}
\end{equation}
where $f(\cdot)$ is an evaluation function that scores a given quality dimension of the generated video, and $\mathcal{L}(\cdot,\cdot)$ is the attack loss measuring the quality degradation from the original video $G(X)$ to the adversarial video $G(X')$. We instantiate the attack loss as the score drop $\mathcal{L} = f(G(X)) - f(G(X'))$.
Considering the inherently spatiotemporal nature of video, we instantiate $f$ with two \textbf{attack objectives}: the semantic objective $f_s$ and the temporal objective $f_t$ (Sect.~\ref{sec:attack_metrics}).
The constraint function $g(\cdot,\cdot)$ quantifies the textual similarity between $X$ and $X'$, which is further decomposed into semantic similarity and formal similarity (Sect.~\ref{sec:evaluation}).
The constraint $g(X, X') \geq \epsilon$, with $\epsilon$ specifying the minimum similarity threshold, 
acts as the textual counterpart to the perturbation constraint, inherently bounding the perturbation to ensure imperceptibility.
We denote by $\mathcal{A}: X \rightarrow X'$ the mapping from an original prompt to its adversarial prompt, termed the \textbf{attack method}, and $G$ represents the black-box T2V models under attack. An overview of the attack pipeline is illustrated in Figure \ref{fig:pipe}.

\subsection{Attack Objective}
\label{sec:attack_metrics}
The attack objectives instantiate the evaluation function $f(\cdot)$ that assesses the quality of a generated video $V$. Grounded in the two principal evaluation dimensions of video generation~\cite{huang2024vbench}, we define a semantic objective and a temporal objective, which jointly guide the optimization and assess the attack performance from complementary perspectives.

\subsubsection{Semantic Objective}
Since T2V models are expected to generate videos that are semantically aligned with the input prompt, preserving this alignment is essential for generated content to faithfully reflect the user’s intent.
We define the semantic objective $f_s$ to quantify the degree of alignment between the original prompt and the generated video.
Formally, $f_s$ is computed in a joint vision-language embedding space as follows:
\begin{equation}
\label{equ:video_text}
f_s(V) = sim\big(E_t(X), E_v(V)\big),
\end{equation}
where $E_t$ and $E_v$ represent the text encoder and video encoder from ViCLIP~\cite{wang2023internvid}, a model demonstrating strong zero-shot performance in video-text retrieval tasks, with the original prompt $X$ serving as the fixed semantic anchor.
We employ cosine similarity $sim(\cdot,\cdot)$ to measure the semantic alignment between the textual features of the prompt and the visual features of the video.

Under the formulation in Eq.~\eqref{equ:formulation}, the semantic attack maximizes the drop of $f_s$, forcing the generated video to diverge from the intended meaning even when the prompt perturbation is imperceptibly small.
Notably, $f_s$ targets the holistic video representation by measuring the alignment between the prompt and the entire generated video, rather than solely focusing on text embeddings in the encoder space. Such a design paradigm is essential, since attacking the text encoder alone exerts limited impact on the final generated video.

\subsubsection{Temporal Objective}
Temporal dynamics is the defining property that distinguishes video generation from image generation, as T2V models must synthesize natural and coherent motion beyond per-frame fidelity.
Generated videos may appear visually plausible frame by frame and yet exhibit unnatural or stagnant motion, a failure mode that the semantic objective alone cannot fully capture.
Therefore, the temporal dimension must be explicitly considered when assessing the adversarial robustness of T2V models.

To this end, we define the temporal objective $f_t$ to quantify the degree of motion dynamics in the generated content. Among the various attributes of the temporal dimension (e.g., motion smoothness, temporal flickering, and dynamic degree), we instantiate $f_t$ as the dynamic degree, which provides a direct indication of whether the motion specified by the prompt is faithfully synthesized. Given that optical flow is the most fundamental technique for motion estimation, we adopt it to quantify the inter-frame temporal variations:
\begin{equation}
f_t(V) = \sum_{\tau=1}^{T-1}\rho\big(V_\tau,V_{\tau+1}\big),
\label{equ:temporal}
\end{equation}
where $\rho(\cdot, \cdot)$ calculates the optical flow intensity between neighboring frames. We utilize a learnable optical flow estimator~\cite{teed2020raft} to ensure differentiability of the entire process. 
The temporal attack maximizes the drop of $f_t$, suppressing the motion specified by the prompt and degrading the temporal realism of the generated video, with a completely static video constituting the extreme case of a successful attack.

\subsection{Attack Method}
\label{sec:attack_methods}
The goal of our attack methods is to construct an adversarial prompt $X'$ that maximizes the attack loss $\mathcal{L}$, \ie, that minimizes the semantic or temporal score of the generated video.
The perturbation applied to the original prompt $X$ is kept minimal, so that $X'$ preserves both semantic and formal similarity to $X$.
Among the four fundamental text editing operations, \ie, substitution, reordering, insertion, and deletion, reordering yields marginal attack effectiveness.
Similarly, deleting non-critical words produces negligible impact, while the deletion of important words violates the minimal perturbation constraint.
Therefore, we focus on two word-level adversarial attack strategies: the substitution attack (T2VAttack-S) and the insertion attack (T2VAttack-I). Both strategies are designed to ensure that adversarial prompts remain semantically consistent and formally natural, while effectively degrading the semantic fidelity or temporal dynamics of the generated videos.

\subsubsection{T2VAttack-S}
Inspired by prominent adversarial attack methods in text classification~\cite{jin2020bert}, we extend the substitution operation to the T2V generation task. Specifically, our method first identifies semantically or temporally critical key words in the prompt $X$, and then replaces them with high-similarity synonyms. This substitution aims to induce significant changes in the generated video content while keeping the modifications imperceptible.
While conventional text classification methods derive attack scores from label confidence, our method computes them based on their effect on semantic or temporal objectives, with similar adjustments applied to word importance and stopping criteria.
To ensure attack stealthiness, we also employ SentenceTransformer~\cite{reimers2020making} for accurate textual similarity measurement and enforce a high similarity threshold.

Formally, given a prompt $X = [\,x_1, x_2, \ldots, x_n\,]$ consisting of $n$ words, we define the importance score $s_i$ for the word $x_i$ as the change in the attack objective $f$ before and after removing $x_i$:
\begin{equation}
s_i = f(G(X)) - f(G(X_{\setminus x_i})),
\label{equ:importance}
\end{equation}
where $X_{\setminus x_i}$ denotes the prompt with $x_i$ removed, and $f$ corresponds to either semantic objective $f_s$ or temporal objective $f_t$ introduced in Sect.~\ref{sec:attack_metrics}.
By prioritizing key words with the highest importance scores, our method enables subtle yet effective modifications that degrade semantic fidelity or temporal dynamics while preserving semantic and formal similarity.

For each important word $x_i$, we retrieve a set of synonym candidates from lexical databases such as WordNet \cite{miller1995wordnet}. To ensure semantic and syntactic coherence in the modified prompt, we adopt a three-step filtering strategy: (1) filter out stop words, (2) select high-similarity synonyms whose similarity exceeds a threshold $\theta$ with $x_i$, measured by SentenceTransformer, and (3) check Part-of-Speech (POS) consistency to ensure grammatical correctness. 
After obtaining valid candidates, we craft a set of perturbed prompts $X'$ by substituting $x_i$ with each candidate and then evaluate their attack objective $f$. 
From the candidates that successfully reach an attack threshold $\tau$ ($\tau_s$ for the semantic criterion and $\tau_t$ for the temporal criterion), we select the substitution with the highest textual similarity to $X$ to maximize stealthiness.
Otherwise, if no candidate meets the threshold, the algorithm proceeds to the next most important word and repeats the previous substitution process.
The entire procedure is conducted in a greedy manner and terminates when either the cumulative number of substitutions reaches the predefined word modification limit or the overall attack objective is achieved, thereby controlling the overall perturbation magnitude.

\subsubsection{T2VAttack-I}
\label{sec:wordaddition}
In contrast to T2VAttack-S, which perturbs existing words in the original prompt $X$, T2VAttack-I introduces adversarial perturbations by prepending additional words, yielding the adversarial prompt $X'$. This approach subtly shifts the semantics without modifying the core syntactic structure of $X$. Specifically, we first randomly sample a set of candidate words $\mathbb{W}^{(1)}=\{w_1,w_2,\cdots,w_{q_1}\}$ from a vocabulary (e.g., GloVe~\cite{pennington2014glove}) as potential first-level prefixes. Each candidate word $w_i \in \mathbb{W}^{(1)}$ is prepended to $X$, forming an adversarial prompt ${X'}^{(1)}_{i} = [\,w_{i};\, x_{1:n}\,]$.
We then evaluate the attack objective $f$ for each perturbed prompt and retain the top $k_1$ candidates that induce the largest semantic or temporal disruption as the first-level prefixes,
\begin{equation}
    \tilde{\mathbb{W}}^{(1)} = \operatorname{TOP-K}_{w_{i} \in \mathbb{W}^{(1)}}^{k_1}
    \left( f\left(G\left(\left[w_{i}; x_{1:n}\right]\right)\right) \right),
\end{equation}
where the $\operatorname{\textbf{TOP-K}}$ operator selects the $k_1$ elements from $\mathbb{W}^{(1)}$ that minimize the objective function $f(\cdot)$, identifying the most effective prefixes.

Based on each retained first-level prefix in $\tilde{\mathbb{W}}^{(1)}$, we further sample $q_2$ additional candidate words as the second-level prefixes and repeat the evaluation and selection process. 
This multi-stage strategy iteratively prepends words, progressively expanding the incremental semantic deviation and amplifying adversarial influence in the generated video. 
The total number of queries in the whole process is $Q = q_1 + k_1 \cdot q_2 + \ldots + k_{i-1} \cdot q_i $. Despite its simplicity, our experiments demonstrate that inserting even a single word with dozens of queries can induce significant deviations in the generated video content.

Although T2VAttack-I typically produces adversarial prompts that remain natural and easily understood by humans, the inserted words may also be more noticeable to detectors. To enhance the stealthiness, we propose a variant, \mbox{T2VAttack-I++}, which introduces character-level perturbations on the inserted prefixes. These character-level modifications include insertion, deletion, reordering, substitution, duplication, symbol replacement, and case flipping. 
These perturbations reduce the readability of the inserted prefixes while keeping the words in the original prompt unchanged, thereby making the modifications harder to detect by human detectors and further compromising T2V generation quality.

\section{Experiments}
\label{sec:exp}

\subsection{Dataset}

Effective adversarial robustness evaluations require both powerful victim models and well-curated prompts. Although VBench \cite{huang2024vbench} provides a comprehensive benchmark for evaluating video generative models, its results reveal that current models still fall short of producing fully complex and coherent videos. This limitation makes direct use of VBench suboptimal for adversarial evaluation, as low performance may obscure the true impact of attacks.
To address this, we build \textbf{T2VAttackBench}, a high-quality textual dataset derived from VBench and refined with large language models like GPT-4o.
Empirical evaluations with both semantic and temporal objectives, as shown in Table \ref{tab:st_metrics}, demonstrate that T2VAttackBench yields videos with better visual-language alignment and motion dynamics compared to the original VBench. The construction process for our dataset is as follows.

\begin{table}[b]
    \centering
    \caption{Comparison of different prompt datasets on state-of-the-art T2V models under semantic and temporal metrics. Our proposed T2VAttackBench yields better alignment and dynamics than VBench \cite{huang2024vbench} and GPT-4o-generated prompts (GPT-G).
    }
    \label{tab:st_metrics}
    \begin{tabular}{c|c|ccc}
        \toprule
        \textbf{Objective} & \textbf{Model} & \textbf{VBench} & \textbf{GPT-G} & \textbf{Ours} \\
        \midrule
        \multirow{4}{*}{{\begin{tabular}[c]{@{}c@{}}Semantic\\ Objective\end{tabular}}} 
        & ModelScope   & 72.6 & 72.2 & 81.1 \\
        & CogVideoX    & 65.5 & 71.4 & 79.8 \\
        & Open-Sora    & 68.4 & 70.7 & 80.1 \\
        & HunyuanVideo & 73.6 & 76.8 & 83.0 \\
        \midrule
        \multirow{4}{*}{{\begin{tabular}[c]{@{}c@{}}Temporal\\ Objective\end{tabular}}} 
        & ModelScope   & 34.0 & 20.8 & 45.2 \\
        & CogVideoX    & 46.8 & 45.8 & 83.6 \\
        & Open-Sora    & 47.7 & 24.8 & 67.9 \\
        & HunyuanVideo & 66.3 & 48.5 & 72.1 \\
        \bottomrule
    \end{tabular}
\end{table}

Prompt Expansion: We augment the VBench dataset by generating semantically and temporally diverse prompts using GPT-4o refinement. This process generates 200 additional textual prompts (denoted GPT-G) covering semantic or temporal dimensions.

Prompt Filtering: To ensure high-quality adversarial samples, we employ a two-step protocol, \ie, Model-Centric Selection and Consensus Filtering. During Model-Centric Selection, four state-of-the-art T2V models independently rank prompts from VBench and GPT-G, retaining the top-$k$ samples with the highest video-text similarity and temporal dynamics scores. For Consensus Filtering, we perform the intersection of the four model-selected subsets. This rigorous process guarantees that prompts in T2VAttackBench: (1) produce videos with strong semantic consistency and dynamic fidelity under normal conditions, and (2) provide a “clean slate” for isolating degradation caused by attacks rather than exposing inherent weaknesses of the models. Detailed specifications, prompt templates, and the complete dataset are provided in Appendix~F.

\begin{table*}[htbp]
    \centering
    \caption{Attack performance on T2V models. We compare our methods (T2VAttack-S, T2VAttack-I, and T2VAttack-I++) against the adapted QueryFreeAdv~\cite{zhuang2023pilot} (a representative T2I attack adapted to T2V models) across four victim models (ModelScope, CogVideoX, Open-Sora, and HunyuanVideo). For QueryFreeAdv, we report the maximum efficacy among its three optimization variants, with their averaged semantic and formal similarities. Metrics include the original and adversarial objective scores, attack efficacy (semantic or temporal difference), semantic similarity, and formal similarity. Semantic Similarity measures the semantic closeness between the original and adversarial prompts (i.e., stealthiness), whereas Semantic Difference reports the post-attack change of the video-text objective (i.e., attack efficacy). \underline{Underlined} marks low adversarial scores; \textbf{Bold} indicates effective attacks per model.
    }
    \begin{tabular}{cc|cc|cc|c}
        \toprule
        \rowcolor{gray!45}
        \textbf{\begin{tabular}[c]{@{}c@{}}Victim\\ Models\end{tabular}} & 
        \textbf{\begin{tabular}[c]{@{}c@{}}Attack\\ Method\end{tabular}} & 
        \textbf{\begin{tabular}[c]{@{}c@{}}Semantic\\ Similarity $\uparrow$\end{tabular}} & 
        \textbf{\begin{tabular}[c]{@{}c@{}}Formal\\ Similarity $\uparrow$\end{tabular}} & 
        \textbf{\begin{tabular}[c]{@{}c@{}}Original\\ Score\end{tabular}} & 
        \textbf{\begin{tabular}[c]{@{}c@{}}Adversarial\\ Score\end{tabular}} & 
        \textbf{Difference $\uparrow$} \\
        \midrule
        \rowcolor{gray!20}
        \multicolumn{6}{c|}{\textbf{Semantic  Attack}} & \multicolumn{1}{c}{\textbf{Semantic Difference}} \\
        \midrule
        \multirow{4}{*}{ModelScope} 
        & QueryFreeAdv~\cite{zhuang2023pilot} & 0.93 & 0.84 & 81.1 & 77.8 & 3.3 \\
        & T2VAttack-S & 0.81 & 0.78 & 81.1 & 56.7 & 24.4 \\
        & T2VAttack-I & 0.89 & 0.91 & 81.1 & 56.0 & \textbf{25.1} \\
        & T2VAttack-I++ & 0.89 & 0.91 & 81.1 & 61.3 & 19.8 \\
        \midrule
        \multirow{4}{*}{CogVideoX} 
        & QueryFreeAdv~\cite{zhuang2023pilot} & 0.93 & 0.81 & 79.8 & 77.0 & 2.8 \\
        & T2VAttack-S & 0.84 & 0.79 & 79.8 & \underline{52.7} & \textbf{27.1} \\
        & T2VAttack-I & 0.90 & 0.91 & 79.8 & 55.1 & 24.7 \\
        & T2VAttack-I++ & 0.90 & 0.91 & 79.8 & 62.3 & 17.5 \\
        \midrule
        \multirow{4}{*}{Open-Sora} 
        & QueryFreeAdv~\cite{zhuang2023pilot}  & 0.93 & 0.81 & 80.1 & 77.7 & 2.4 \\
        & T2VAttack-S  & 0.82 & 0.78 & 80.1 & 54.5 & 25.6 \\
        & T2VAttack-I  & 0.90 & 0.91 & 80.1 & \underline{50.0} & \textbf{30.1} \\
        & T2VAttack-I++  & 0.90 & 0.91 & 80.1 & 60.4 & 19.7 \\
        \midrule
        \multirow{4}{*}{HunyuanVideo} 
        & QueryFreeAdv~\cite{zhuang2023pilot} & 0.92 & 0.83 & 83.0 & 82.2 & 0.8 \\
        & T2VAttack-S & 0.82 & 0.78 & 83.0 & 58.8 & \textbf{24.2} \\
        & T2VAttack-I & 0.89 & 0.91 & 83.0 & 61.6 & 21.4 \\
        & T2VAttack-I++ & 0.89 & 0.91 & 83.0 & 67.1 & 15.9 \\
        \midrule
        \midrule
        \rowcolor{gray!20}
        \multicolumn{6}{c|}{\textbf{Temporal  Attack}} & \multicolumn{1}{c}{\textbf{Temporal Difference}} \\
        \midrule
        \multirow{4}{*}{ModelScope} 
        & QueryFreeAdv~\cite{zhuang2023pilot} & 0.87 & 0.77 & 45.2 & 39.8 & 5.4 \\
        & T2VAttack-S & 0.87 & 0.80 & 45.2 & 9.3 & 35.9 \\
        & T2VAttack-I & 0.83 & 0.86 & 45.2 & \underline{3.4} & \textbf{41.8} \\
        & T2VAttack-I++ & 0.83 & 0.86 & 45.2 & 16.9 & 28.3 \\
        \midrule
        \multirow{4}{*}{CogVideoX} 
        & QueryFreeAdv~\cite{zhuang2023pilot} & 0.87 & 0.70 & 83.6 & 76.9 & 6.7 \\
        & T2VAttack-S & 0.77 & 0.75 & 83.6 & 29.3 & 54.3 \\
        & T2VAttack-I & 0.82 & 0.86 & 83.6 & 20.7 & \textbf{62.9} \\
        & T2VAttack-I++ & 0.82 & 0.86 & 83.6 & 33.3 & 50.3 \\
        \midrule
        \multirow{4}{*}{Open-Sora} 
        & QueryFreeAdv~\cite{zhuang2023pilot}  & 0.87 & 0.70 & 67.9 & 69.3 & $-$1.4 \\
        & T2VAttack-S  & 0.82 & 0.77 & 67.9 & 19.4 & 48.5 \\
        & T2VAttack-I  & 0.82 & 0.86 & 67.9 & \underline{5.5} & \textbf{62.4} \\
        & T2VAttack-I++  & 0.83 & 0.86 & 67.9 & 19.2 & 48.7 \\
        \midrule
        \multirow{4}{*}{HunyuanVideo} 
        & QueryFreeAdv~\cite{zhuang2023pilot} & 0.86 & 0.72 & 72.1 & 73.5 & $-$1.4 \\
        & T2VAttack-S & 0.69 & 0.64 & 72.1 & 36.6 & \textbf{35.5} \\
        & T2VAttack-I & 0.84 & 0.86 & 72.1 & 42.2 & 29.9 \\
        & T2VAttack-I++ & 0.85 & 0.86 & 72.1 & 50.0 & 22.1 \\
        \bottomrule
    \end{tabular}
    \label{tab:attackers_victims}
    \vspace{-10pt}
\end{table*}

\subsection{Evaluation Metric}
\label{sec:evaluation}

To comprehensively evaluate the adversarial robustness of T2V models, we assess adversarial prompts along three dimensions: stealthiness (instantiating the constraint function $g(\cdot,\cdot)$), efficacy (reporting the realized attack loss $\mathcal{L}$), and efficiency.

For \textbf{adversarial prompt stealthiness}, we consider semantic similarity, formal similarity, and word modification count. The \textbf{semantic similarity} quantifies the degree to which an adversarial prompt $X'$ preserves the semantic meaning of the original prompt $X$, computed as the cosine similarity between SentenceTransformer~\cite{reimers2020making} embeddings. 
The \textbf{formal similarity} reflects the structural similarity between $X$ and $X'$, quantified via normalized Levenshtein distance. 
The \textbf{word modification count} denotes the number of word-level edits applied to $X$.

For \textbf{attack efficacy}, we report the \textbf{Original Score} $f(G(X))$ and the \textbf{Adversarial Score} $f(G(X'))$, representing the objective scores of the videos generated from the original and adversarial prompts, respectively.
We then measure their difference $\Delta = f(G(X)) - f(G(X'))$, where $f$ corresponds to either the semantic or the temporal attack objective. We refer to these differences as the \textbf{Semantic Difference $\Delta_s$} and the \textbf{Temporal Difference $\Delta_t$}.

For \textbf{attack efficiency}, we report the total number of \textbf{queries} to the black-box T2V model, where each query corresponds to a complete video generation process.

\subsection{Main Results}

We conduct comprehensive experiments on four state-of-the-art T2V diffusion models, including ModelScope, CogVideoX, Open-Sora, and HunyuanVideo. Table~\ref{tab:attackers_victims} reports the performance of our proposed methods (T2VAttack-S, T2VAttack-I, and \mbox{T2VAttack-I++}) alongside the adapted QueryFreeAdv~\cite{zhuang2023pilot} under both semantic and temporal objectives. Since no existing adversarial framework directly targets T2V generation, we adapt QueryFreeAdv to target the specific text encoder employed by each victim model (\ie, CLIP for ModelScope, T5 for CogVideoX and Open-Sora, and Llama for HunyuanVideo) using its official five-character perturbation protocol. Notably, the fragmentation of these random characters during subword tokenization substantially degrades token-level formal similarity.
We report the strongest results across its three optimization variants (greedy search, genetic algorithm, and PGD~\cite{madry2018towards}), with complete configurations detailed in Appendix~B.

Several key observations emerge from these evaluations. Both semantic and temporal attacks effectively disrupt all four T2V models. The models exhibit distinct vulnerability patterns: CogVideoX and Open-Sora show higher sensitivity to semantic attacks, ModelScope and Open-Sora are more vulnerable to temporal attacks, whereas HunyuanVideo demonstrates notably stronger robustness across all attack types. Specifically, T2VAttack-S reduces the semantic score of CogVideoX by 27.1 points (34.0\% drop), while T2VAttack-I decreases that of Open-Sora by 30.1 points (37.6\% drop). Temporally, T2VAttack-I drives the dynamics scores of ModelScope and Open-Sora down to near-zero (3.4 and 5.5, corresponding to 92.5\% and 91.9\% reductions), almost completely suppressing their motion. While T2VAttack-I generally outperforms T2VAttack-S under the temporal objective, HunyuanVideo's prompt rewrite mechanism~\cite{sun2024hunyuan} effectively neutralizes this insertion strategy. Although T2VAttack-I++ restores visual stealthiness, it introduces a trade-off by marginally compromising attack efficacy.

In stark contrast, the adapted QueryFreeAdv~\cite{zhuang2023pilot} exerts negligible impact on T2V generation. The strongest variant reduces the semantic score by at most 3.3 points. Temporally, it fails to suppress motion and even inadvertently increases the dynamic degree of videos generated by Open-Sora and HunyuanVideo. This inefficiency stems from a fundamental decoupling between the text-encoder space and video-level semantics. Although QueryFreeAdv successfully induces significant text-embedding displacements, these geometric shifts exhibit a near-zero rank correlation with the actual drop in video-text consistency (please refer to Appendix~B for further implementation details and analysis). This inherent feature decoupling precisely explains the failure of T2I attacks and validates the necessity of the video-level optimization objectives adopted in T2VAttack.

\begin{table*}[!t]
    \centering
    \caption{Ablation studies on T2VAttack-I strategy under varying conditions: (a) different vocabulary sources, (b) varying word modification count and query allocations, (c) word insertion positions. \colorbox{gray!20}{gray} is the default setting.
    }
    \vspace{-10pt}
    \label{tab:combined_analysis}

    \subfloat{
        \begin{minipage}[t]{0.40\textwidth}
            \centering
            {\small (a) Vocabulary}
            \par\vspace{8pt}
            \resizebox{\linewidth}{!}
            {\renewcommand{\arraystretch}{1.1}
            \begin{tabular}{lccc}
                \toprule
                \textbf{\makecell{Vocabulary\\ Source}} & \textbf{\makecell{Original\\ Size}} & \textbf{\makecell{Filtered\\ Size}} & \textbf{\makecell{Semantic\\ Difference}} \\
                \midrule
                CLIP~\cite{radford2021learning}        & 49,408    & 18,557    & 23.1 \\
                T5~\cite{t5}          & 32,000    & 11,188    & 22.2 \\
                LLaVA~\cite{liu2023visual}       & 128,000   & 5,377     & 21.3 \\
                \cmidrule(lr){1-4}
                BERT~\cite{devlin2019bert}        & 30,522    & 16,959    & 24.5 \\
                GPT2~\cite{radford2019language}         & 50,257    & 3,641     & 20.6 \\
                \cmidrule(lr){1-4}
                Word2Vec~\cite{mikolov2013efficient}    & 3,000,000 & 80,313    & 23.4 \\
                \rowcolor{gray!20}
                GloVe~\cite{pennington2014glove} & 400,001   & 74,495    & \textbf{25.1} \\
                \bottomrule
            \end{tabular}
            }
        \end{minipage}
    }
    \hfill 
    \subfloat{
        \begin{minipage}[t]{0.29\textwidth} 
            \centering
            {\small (b) Word Edits}
            \par\vspace{8pt}
            \resizebox{\linewidth}{!}{
            \begin{tabular}{lccl} 
                \toprule
                \textbf{\makecell{Words\\ Edits}} & 
                \textbf{\makecell{Query\\ Count}} & 
                \multicolumn{2}{c}{\textbf{\makecell{Semantic\\ Difference}}} \\ 
                \midrule
                \multirow{4}{*}{\makecell{1}} 
                & 8 & \multicolumn{2}{c}{11.9} \\
                & 16 & \multicolumn{2}{c}{16.7} \\
                & 32 & \multicolumn{2}{c}{20.3} \\
                \rowcolor{gray!20}
                & 64 & \multicolumn{2}{c}{\textbf{25.1}} \\
                \cmidrule(lr){1-4}
                \multirow{4}{*}{\makecell{2}} 
                & $4 + 1 \times 60$ & \multicolumn{2}{c}{28.6} \\
                & $4 + 4 \times 15$ & \multicolumn{2}{c}{26.4} \\
                & $16 + 1 \times 48$ & \multicolumn{2}{c}{\textbf{30.1}} \\
                & $16 + 4 \times 12$ & \multicolumn{2}{c}{29.8} \\
                \bottomrule
            \end{tabular}
            }
        \end{minipage}
    }
    \hfill 
    \subfloat{
        \begin{minipage}[t]{0.24\textwidth}
            \centering
            {\small (c) Position}
            \par\vspace{8pt}
            \resizebox{\linewidth}{!}{
            \renewcommand{\arraystretch}{1.2}
            \begin{tabular}{lc} 
                \toprule
                \textbf{\makecell{Insertion\\ Position}} & \textbf{\makecell{Semantic\\ Difference}} \\
                \midrule
                \rowcolor{gray!20}
                First     & \textbf{25.1} \\
                Middle    & 21.6 \\
                Last      & 20.0 \\
                Random    & 21.1 \\
                Important & 23.3 \\
                \bottomrule
            \end{tabular}
            }
        \end{minipage}
    }
\end{table*}

\begin{table*}[htbp]
    \centering
    \caption{Ablation studies on T2VAttack-S strategy under varying conditions: (a) varying word modification count and query allocations, (b) synonym filtering threshold $\theta$, (c) attack score criterion $\tau_s$ on semantic objective, (d) attack score criterion $\tau_t$ on temporal objective.
    \colorbox{gray!20}{gray} is the default setting.
    }
    \vspace{-8pt}
    \label{tab:comprehensive_attack_analysis}
    \subfloat{
        \begin{minipage}[t]{0.26\textwidth}
            \centering
            {\renewcommand{\arraystretch}{1.3}
            {\small (a) Word Edits}
            \par\vspace{5pt}
            \resizebox{\linewidth}{!}{
            \begin{tabular}{ccc}
                \toprule
                \textbf{\makecell{Words\\ Edits}} & \textbf{\makecell{Query\\ Count}} & \textbf{\makecell{Semantic\\ Difference}} \\
                \midrule
                1 & 22.3 & 16.1 \\
                2 & 28.4 & 20.5 \\
                \rowcolor{gray!20}
                3 & 34.3 & \textbf{24.4} \\
                \bottomrule
            \end{tabular}
            }
            }
        \end{minipage}
    }
    \hfill
    \subfloat{
        \begin{minipage}[t]{0.22\textwidth}
            \centering
            {\small (b) Synonym Threshold $\theta$}
            \par\vspace{5pt}
            \resizebox{\linewidth}{!}{
            \begin{tabular}{cc}
                \toprule
                \textbf{\makecell{Synonym \\ Threshold}} & \textbf{\makecell{Semantic\\ Difference}} \\
                \midrule
                0.90 & 24.2 \\
                \rowcolor{gray!20}
                0.80 & 24.4 \\
                0.70 & 25.1 \\
                0.60 & 25.2 \\
                0.50 & \textbf{25.5} \\
                \bottomrule
            \end{tabular}
            }
        \end{minipage}
    }
    \hfill
    \subfloat{
        \begin{minipage}[t]{0.22\textwidth}
            \centering
            {\small (c) Semantic Criterion $\tau_s$}
            \par\vspace{5pt}
            \resizebox{\linewidth}{!}{
            \begin{tabular}{cc}
                \toprule
                \textbf{\makecell{Semantic\\Criterion}} & \textbf{\makecell{Semantic\\ Difference}} \\
                \midrule
                0.30 & \textbf{29.4} \\
                0.40 & 27.8 \\
                \rowcolor{gray!20}
                0.50 & 24.4 \\
                0.60 & 21.6 \\
                0.70 & 17.4 \\
                \bottomrule
            \end{tabular}
            }
        \end{minipage}
    }
    \hfill
    \subfloat{
        \begin{minipage}[t]{0.22\textwidth}
            \centering
            {\small (d) Temporal Criterion $\tau_t$}
            \par\vspace{5pt}
            \resizebox{\linewidth}{!}{
            \begin{tabular}{cc}
                \toprule
                \textbf{\makecell{Temporal\\Criterion}} & \textbf{\makecell{Temporal\\ Difference}} \\
                \midrule
                0.02 & \textbf{37.3} \\
                0.05 & 36.6 \\
                \rowcolor{gray!20}
                0.10 & 35.9 \\
                0.15 & 35.3 \\
                0.20 & 34.5 \\
                \bottomrule
            \end{tabular}
            }
        \end{minipage}
    }
\end{table*}

\subsection{Ablation Study}

We perform extensive ablation studies and in-depth analyses of our attack methods. These include parameter ablations for three attack methods: T2VAttack-I, T2VAttack-S, and \mbox{T2VAttack-I++}, as well as three discussions covering: Text Editing Operations, Attack Stealthiness, and Part-of-Speech. All of these experiments are conducted on ModelScope.

\subsubsection{T2VAttack-I Ablations}

We first investigate how vocabulary selection affects the performance of the T2VAttack-I strategy in Table~\ref{tab:combined_analysis}-(a).
Candidate vocabularies are derived from T2V-related text encoder (e.g., CLIP~\cite{radford2021learning}, T5~\cite{t5}, LLaVA~\cite{liu2023visual}), traditional NLP embeddings (Word2Vec~\cite{mikolov2013efficient}, GloVe~\cite{pennington2014glove}), and large language models (BERT~\cite{devlin2019bert}, GPT~\cite{radford2019language}).
Except for LLaVA, which employs the LLaVA-Llama-3-8B variant~\cite{2023xtuner}, all models use their official vocabularies.
Since modern large models commonly employ subword tokenization, we apply a spell-check filter to retain only valid words. The results indicate that GloVe achieves the highest overall attack score of 25.1 with a filtered vocabulary size of 74,495 tokens. Despite a smaller post-filtering vocabulary, BERT retains competitive attack effectiveness. This suggests that carefully selecting a semantically relevant subset of words is crucial for attack efficacy.
Table~\ref{tab:combined_analysis}-(b) analyzes the impact of word modifications and query counts on T2VAttack-I. Increasing the number of queries significantly enhances the semantic objective difference. Extending the attack to two-word insertions further reduces semantic similarity. For instance, allocating $16 + 4 \times 12$ queries reduces the semantic score of the ModelScope model by 29.8. This query allocation involved 16 queries to select the top 4 most influential words for the first prefix, followed by 12 queries for each of these words to select the second prefix, with a total of 64 queries. Notably, inserting a single prefix word with 64 queries generally achieves strong attack performance.
Table~\ref{tab:combined_analysis}-(c) investigates the effect of insertion positions on T2VAttack-I. Surprisingly, inserting words at the beginning of the prompt proves to be the most effective, reducing the semantic score of ModelScope by 25.1. This even outperforms inserting a word at the most important position in the prompt, whose importance is calculated using Equation \ref{equ:importance}. In contrast, inserting a word at the end of the prompt has the least impact.

\subsubsection{T2VAttack-S Ablations}

Table~\ref{tab:comprehensive_attack_analysis} analyzes the impact of word modifications and query allocations, synonym thresholds, and attack score thresholds on T2VAttack-S. The word modification count, which represents the maximum number of substitutable words, controls the query count during the attack process. This query count includes the importance calculation of each word in the prompt when ranking and querying adversarial prompts for synonym substitution. As shown in Table~\ref{tab:comprehensive_attack_analysis}-(a), as the number of substitutable words increases, both the query count and attack effectiveness increase. To maintain the stealthiness of adversarial prompts, we limit the number of word modifications to no more than three.
Table~\ref{tab:comprehensive_attack_analysis}-(b) investigates the impact of the synonym threshold on T2VAttack-S. Since our synonym candidates are sourced from WordNet lexical databases, this candidate set already has a high similarity to the target words. As a result, the overall change in semantic similarity is minimal. However, this synonym filtering process can optimize synonym candidates for some adversarial prompts, thereby improving adversarial prompt stealthiness due to the polysemy inherent in WordNet.
Table~\ref{tab:comprehensive_attack_analysis}-(c) and Table~\ref{tab:comprehensive_attack_analysis}-(d) analyze the influence of attack score thresholds on T2VAttack-S in semantic and temporal attacks. The attack score threshold represents the exit criterion for the substitution process, and its optimal value differs significantly between semantic and temporal objectives. A higher criterion threshold corresponds to a looser exit condition for an attack, leading to lower attack effectiveness but improved efficiency.
For simple examples, a low criterion threshold struggles to achieve high stealthiness and high-quality adversarial examples; for complex examples, a high criterion threshold makes it difficult to generate effective adversarial examples. Therefore, the criterion threshold represents a delicate trade-off between stealthiness and effectiveness in the T2VAttack-S strategy.

\begin{table}[t]
    \centering
    \caption{Ablation studies of character-level perturbation count on \mbox{T2VAttack-I++} strategy.
    '-' denotes T2VAttack-I baseline results.
    \colorbox{gray!20}{gray} is the default setting.
    }
    \begin{tabular}{c|c}
        \toprule
        \textbf{\begin{tabular}[c]{@{}c@{}}Char-perturbations\\ Query Count\end{tabular}} & \textbf{\makecell{Semantic\\ Difference}} \\
        \midrule
        -    & 25.1 \\
        8    & 16.8 \\
        \rowcolor{gray!20}
        16   & 19.8 \\
        24   & 21.1 \\
        32   & 23.3 \\
        \bottomrule
    \end{tabular}
    \label{tab:char_perturbation}
\end{table}

\begin{table}[t]
    \centering
    \caption{Comparison of varying text editing operations (substitution, reordering, insertion, and deletion) at different word positions (First and Important).
    }
    \renewcommand{\arraystretch}{1.1}
    \begin{tabular}{cc|cc|c} 
        \toprule
        \textbf{\begin{tabular}[c]{@{}c@{}}Text\\ Editing\end{tabular}} &
        \textbf{Position} &
        \textbf{\begin{tabular}[c]{@{}c@{}}Semantic\\ Similarity $\uparrow$\end{tabular}} &
        \textbf{\begin{tabular}[c]{@{}c@{}}Formal\\ Similarity $\uparrow$\end{tabular}} &
        \textbf{\makecell{Semantic\\ Difference $\uparrow$}} \\
        \midrule
        \multirow{2}{*}{Insertion}
        & First & 0.89 & 0.91 & 25.1 \\
        & Important & 0.93 & 0.91 & 23.3 \\
        \cline{1-5}
        \multirow{2}{*}{Substitution}
        & First & 0.89 & 0.91 & 4.0 \\
        & Important & 0.82 & 0.91 & 21.6 \\
        \cline{1-5}
        \multirow{2}{*}{Deletion}
        & First & 0.97 & 0.91 & 2.4 \\
        & Important & 0.89 & 0.91 & 21.8 \\
        \cline{1-5}
        \multirow{2}{*}{Reordering}
        & First & 0.97 & 0.82 & 3.3 \\
        & Important & 0.98 & 0.82 & 7.9 \\
        \bottomrule
    \end{tabular}
    \label{tab:attackers_position}
\end{table}

\subsubsection{\mbox{T2VAttack-I++} Ablations}

Table~\ref{tab:char_perturbation} analyzes the impact of additional character-level perturbations on \mbox{T2VAttack-I++}. We treat the one-word, 64-query T2VAttack-I as the baseline, and then apply random character-level modifications to the inserted prefix words. These modifications, including character insertion, deletion, reordering, substitution, duplication, symbol replacement, and case flipping, obscure the newly added semantics and thereby enhance the attack's stealthiness. 
Consequently, these perturbations initially decrease the semantic score difference due to the semantic disruption. However, as the number of character perturbations increases, the reduction effect gradually diminishes. For attack efficiency, \mbox{T2VAttack-I++} applies 16 additional character-level perturbation queries, yielding a total of 80 queries per attack.

\subsubsection{Text Editing Operations}

In NLP, four fundamental adversarial operations are widely employed in text classification tasks: substitution, reordering, insertion, and deletion \cite{zhang2024text}. We adopt analogous operations to assess the adversarial robustness of T2V models: substituting a word with a random word, reordering two words, inserting a random word, and deleting a word. For the substitution and insertion operations, the newly introduced random words are sourced from the GloVe vocabulary. 

Table~\ref{tab:attackers_position} presents the impact of these varying text editing operations, with editing positions including the first position and the most important word position (determined by Equation \ref{equ:importance}). For example, the Insertion-Important operation inserts a random word before the most important word, while the Reordering-First operation swaps the first word with a randomly selected word. The results yield two key findings: 
First, the insertion operation exerts a significant influence on the semantic score, whereas reordering has a minor effect. Second, substitution and deletion operations substantially reduce the semantic score only when applied to the important word position, with negligible impact at other positions.
Considering that the deletion operation introduces more noticeable modifications to the prompt, particularly when applied to important positions, we employ substitution and insertion as our two primary attack strategies. More experimental results and details for other editing positions are provided in Appendix~C.

\begin{table}[t]
    \centering
    \caption{Comparison of different stealthiness conditions: character-level perturbations and random position insertion strategy.
    }
    \renewcommand{\arraystretch}{1.1}
    \begin{tabular}{cc|cc|c} 
        \toprule
        \multicolumn{2}{c}{\textbf{Stealthiness}} & 
        \multirow{3}{*}{\textbf{\begin{tabular}[c]{@{}c@{}}Semantic\\ Similarity $\uparrow$\end{tabular}}} & 
        \multirow{3}{*}{\textbf{\begin{tabular}[c]{@{}c@{}}Formal\\ Similarity $\uparrow$\end{tabular}}} & 
        \multirow{3}{*}{\textbf{\makecell{Semantic\\ Difference $\uparrow$}}} \\ 
        \cmidrule(lr){1-2} 
        \multirow{2}{*}{\textbf{\begin{tabular}[c]{@{}c@{}}Character\\ Perturbations\end{tabular}}} & 
        \multirow{2}{*}{\textbf{\begin{tabular}[c]{@{}c@{}}Random\\ Insertion\end{tabular}}} &
        & & \\ 
        & & & & \\
        \midrule
        & & 0.89 & 0.91 & 25.1 \\
        $\checkmark$ & & 0.89 & 0.91 & 19.8 \\
        & $\checkmark$ & 0.91 & 0.91 & 21.1 \\
        $\checkmark$ & $\checkmark$ & 0.91 & 0.91 & 17.2 \\
        \bottomrule
    \end{tabular}
    \label{tab:stealthiness_attack}
\end{table}

\begin{figure*}[!ht]
    \centering
    \subfloat{
        \begin{minipage}[b]{0.48\linewidth}
            \centering
            \includegraphics[width=\linewidth]{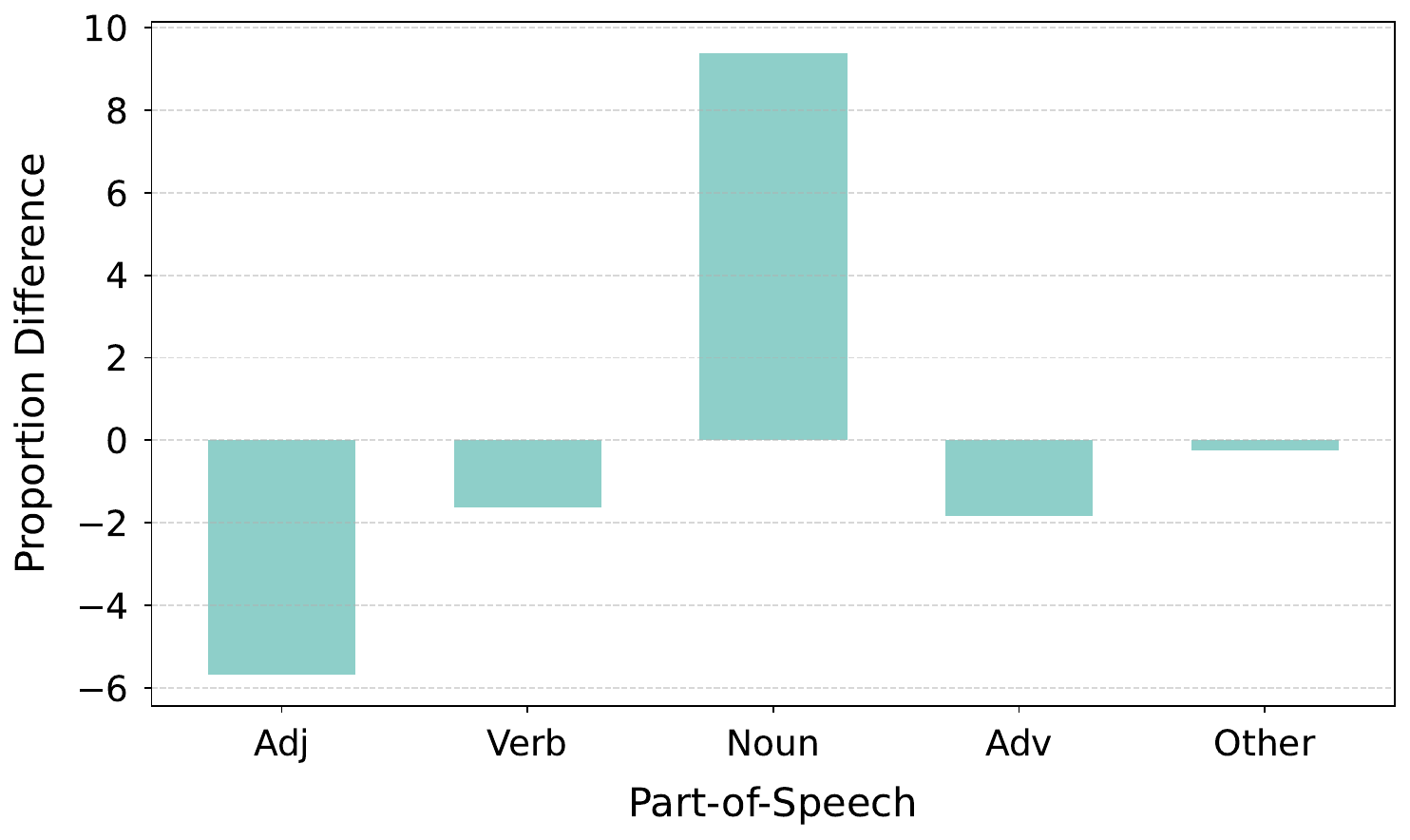}
            \par\vspace{2pt}
            {\small (a) Semantic Attacks}
        \end{minipage}
        \label{fig:vis3_left}
    }
    \hfill
    \subfloat{
        \begin{minipage}[b]{0.48\linewidth}
            \centering
            \includegraphics[width=\linewidth]{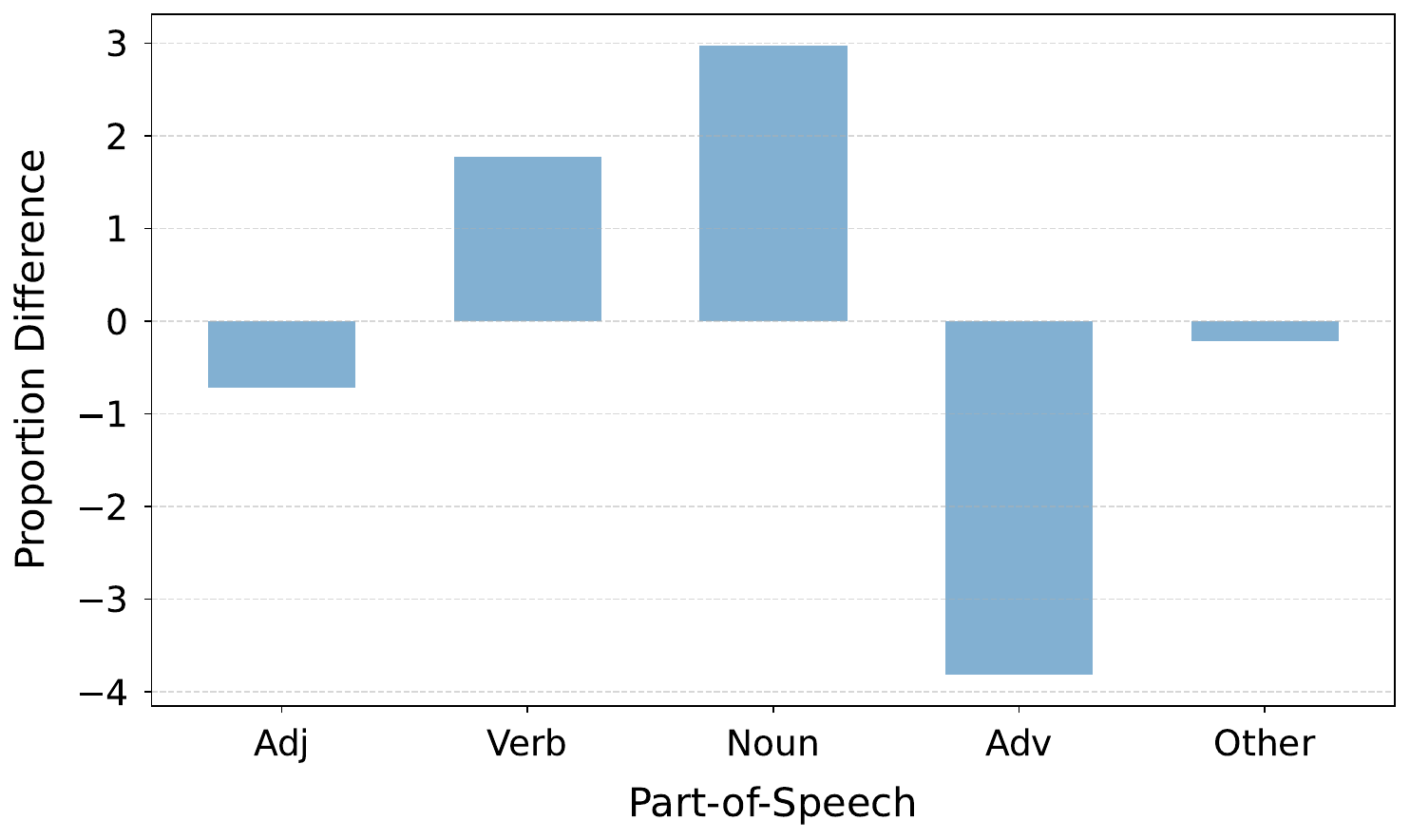}
            \par\vspace{2pt}
            {\small (b) Temporal Attacks}
        \end{minipage}
        \label{fig:vis3_right}
    }
    \caption{Part-of-Speech (POS) analysis on T2VAttack-I. The figure plots the proportion difference between the POS distribution of the most effective adversarial word and the full vocabulary, with a positive value indicating a higher likelihood of a specific POS being an effective adversarial token.
    (a) Semantic attacks: nouns dominate, strongly affecting semantic content.
    (b) Temporal attacks: nouns and verbs dominate, perturbing temporal dynamics.
    }
    \label{fig:pos_effect}
\end{figure*}

\subsubsection{Attack Stealthiness}
For sentiment classification tasks in NLP, replacing “accept” with “reject” can flip the sentiment from positive to negative. In contrast, T2V models exhibit a lower tolerance for such obvious perturbations. For example, replacing “Two pandas discussing an academic paper” with “Two women discussing an academic paper” may not constitute a successful adversarial example. To further enhance the stealthiness of adversarial perturbations, we introduce additional constraints into our attack methods. 

In T2VAttack-S, we restrict the source of substituted words to synonyms and apply a three-step “filter–select–check" constraint mechanism. In T2VAttack-I, we consider two constraints to improve stealthiness: character-level perturbations and random insertion positions. Table~\ref{tab:stealthiness_attack} shows the impact of these constraints on attack effectiveness, by semantic attacks on ModelScope. Character perturbations result in a negative impact of 5.3 in semantic difference, while random insertion position causes a negative impact of 4.0. These results demonstrate a clear trade-off between attack stealthiness and effectiveness. It is worth noting that while character perturbations can obscure the intent of the introduced adversarial words, they are prone to be detected and corrected by specific methods \cite{goyal2023survey}. For human observers, such perturbations improve stealthiness; for machines, they diminish it. Therefore, we retain both T2VAttack-I and \mbox{T2VAttack-I++} to address different environmental conditions.

\subsubsection{Part-of-Speech (POS)}

To further investigate the behavior of the \textit{T2VAttack-I} attack, we analyze the POS distribution of adversarial words under a 64-query, one-word insertion setting. Specifically, we compute the proportion difference between the POS distribution of the selected adversarial word (i.e., the most effective word actually inserted into each prompt) and that of the full candidate vocabulary. A larger difference indicates that words of a particular POS type are more likely to dominate in terms of attack effectiveness.

As illustrated in Fig.~\ref{fig:pos_effect}, for \textbf{semantic attacks}, \textbf{nouns} show the largest increase among the selected adversarial words, suggesting their strong impact on visual-semantic disruption. For \textbf{temporal attacks}, both \textbf{nouns} and \textbf{verbs} contribute significantly, reflecting their critical ability in disturbing motion modeling and temporal consistency. These findings emphasize the importance of POS-aware adversarial strategies and provide insights into linguistic patterns that more effectively degrade T2V model performance.

\begin{figure*}[!ht]
    \centering
    \vspace{10pt}
    \includegraphics[width=0.98\linewidth]{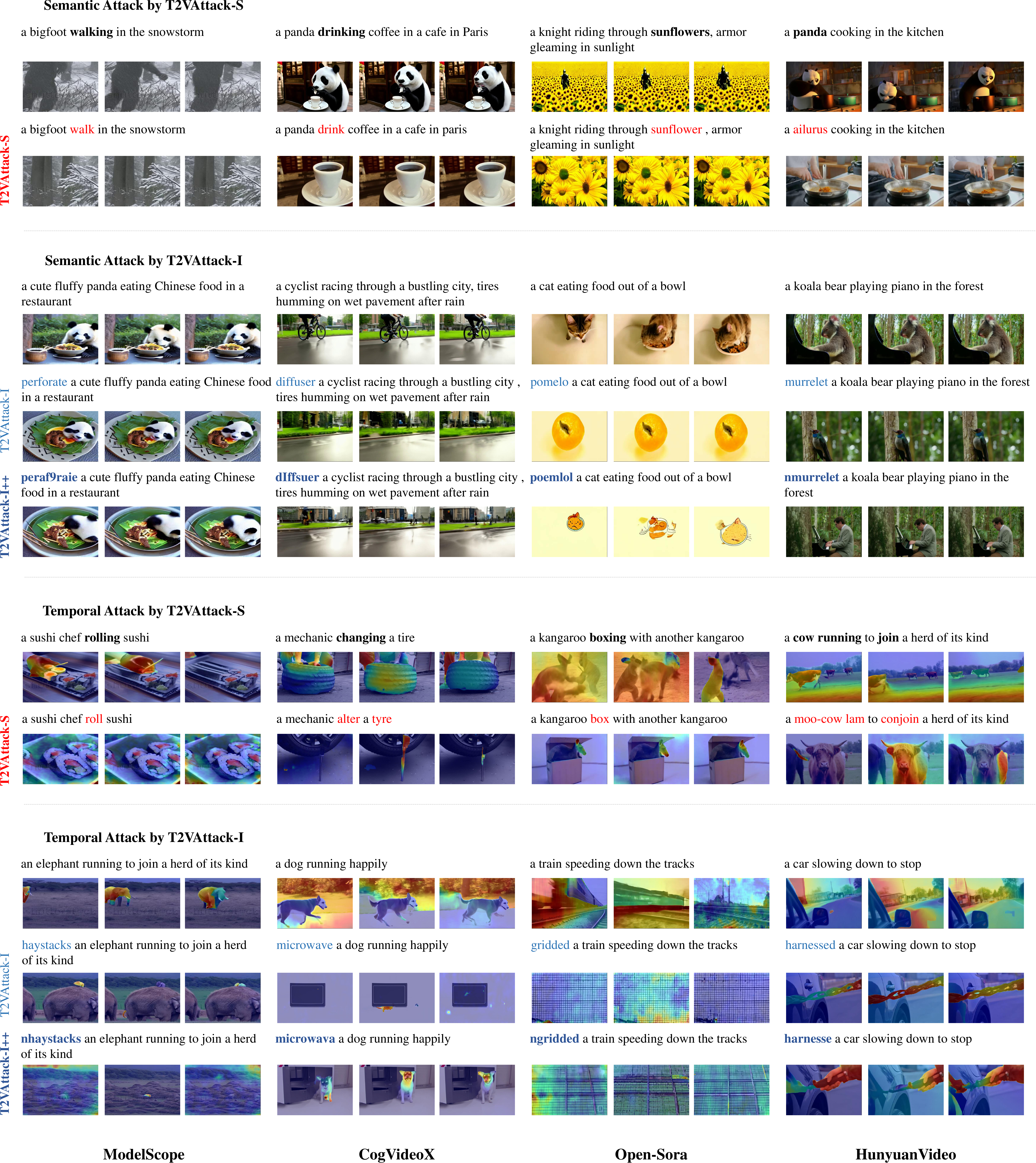}
    \caption{Visualization of adversarial attack effects across four T2V models (columns: ModelScope, CogVideoX, Open-Sora, HunyuanVideo). Rows correspond to: (1) Semantic attack\:-\:T2VAttack-S; (2) Semantic attack\:-\:T2VAttack-I; (3) Temporal attack\:-\:T2VAttack-S; (4) Temporal attack\:-\:T2VAttack-I. In Substitution rows, each example shows a pair (top: video generated from the original prompt; bottom: after \textcolor{red}{T2VAttack-S}). In Insertion rows, each example shows a triplet (top: original; middle: after \textcolor[RGB]{71,130,188}{T2VAttack-I}; bottom: after \textcolor[RGB]{25,50,110}{\mbox{T2VAttack-I++}}). For temporal attack cases, the corresponding optical flow magnitudes are visualized to expose motion degradation. These examples highlight that minor prompt perturbations can induce substantial degradation in semantic fidelity and temporal dynamics.
    }
    \label{fig:visualition}
    \vspace{16pt}
\end{figure*}

\subsection{Visualization}

Fig.~\ref{fig:visualition} provides representative visualization examples illustrating the effects of subtle textual perturbations introduced by T2VAttack-I, T2VAttack-S, and \mbox{T2VAttack-I++} attacks across four T2V models: ModelScope, CogVideoX, Open-Sora, and HunyuanVideo, with temporal attacks illustrated via optical flow magnitudes to highlight motion changes. Two key patterns emerge from these observations. 
First, T2VAttack-S emphasizes local and strict semantic constraints by replacing words with close synonyms. For instance, in CogVideoX,  given the prompt ``a panda drinking coffee in a cafe in Paris'', merely substituting ``drinking'' with ``drink'' induces the complete omission of the central subject (panda) in the video generated, despite the negligible lexical change.
Second, T2VAttack-I, limited to a single word insertion, strategically inserts the word most likely to introduce a conflicting or disruptive concept. For example, prepending ``diffuser'' (or its variant ``d\textbf{\textit{I}}ff\textbf{\textit{su}}er'') to ``a cyclist racing through a bustling city, tires humming on wet pavement after rain'' forces CogVideoX to entirely omit the primary visual subjects (the cyclist and bicycle), collapsing the intended scene.
Collectively, these examples underscore that even minor prompt perturbations can significantly compromise semantic fidelity and temporal consistency in T2V diffusion models.
More challenging visualization results on HunyuanVideo are provided in Appendix~G.

\subsection{Limitation}

\textbf{Computational Cost}: Both T2VAttack-S and T2VAttack-I are score-based black-box attacks. As these methods require multiple queries to the victim model, the total runtime is inherently bottlenecked by both the query count and the model's inference efficiency. T2V models are generally highly resource-intensive and time-consuming, for instance, producing the highest-quality video on HunyuanVideo can take up to 50 minutes per inference. Despite imposing strict limitations on the query budget, the overall attack process remains computationally expensive.

\textbf{Challenges in Temporal Attack}: 
As discussed in Sect.~\ref{sec:attack_metrics}, a completely static video constitutes the extreme case of a successful temporal attack.
For CogVideoX and HunyuanVideo, however, some adversarial prompts only partially diminish the temporal dynamics of the generated videos, while others exploit degenerate solutions (e.g., shrinking the size of moving objects or reducing background motion) rather than suppressing motion itself. Consequently, achieving strong temporal attacks remains an open challenge for future research.

\section{Conclusion}
\label{sec:conclusion}
This paper presents a systematic investigation into the adversarial robustness of Text-to-Video diffusion models. We formalize textual adversarial attacks on T2V generation as a unified optimization problem, instantiate it with semantic and temporal objectives, and propose two black-box attack methods, T2VAttack-S and T2VAttack-I, together with the curated T2VAttackBench benchmark.
Our experimental results reveal that even single-word modifications can induce significant semantic shifts or degraded temporal dynamics in the generated videos. 
Despite the simplicity of our methods, these findings expose critical vulnerabilities in the cross-modal alignment and temporal modeling of current T2V models, underscoring the urgent need to improve their adversarial robustness.
Future work should focus on developing effective countermeasures to mitigate such adversarial threats, thereby ensuring safer and more reliable video generation in real-world applications.

\bibliographystyle{IEEEtran}
\bibliography{main}

\clearpage

\section*{Biography Section}
 
\vspace{11pt}

\begin{IEEEbiography}[{\includegraphics[width=1in,height=1.25in,clip,keepaspectratio]{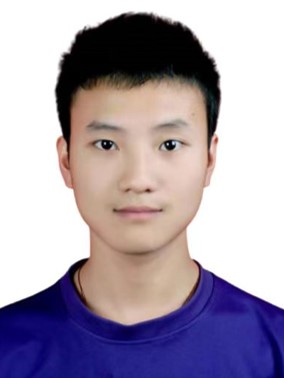}}]{Changzhen Li}
is currently pursuing a Ph.D. degree at the University of Chinese Academy of Sciences (UCAS), Beijing, China. He received his Bachelor's degree (B.S.) from East China Normal University, Shanghai, China, in 2018, and his Master's degree (M.S.) from the Institute of Computing Technology (ICT), Chinese Academy of Sciences (CAS), Beijing, China, in 2022. His research interests include video understanding and the safety of video generation.
\end{IEEEbiography}

\begin{IEEEbiography}[{\includegraphics[width=1in,height=1.25in,clip,keepaspectratio]{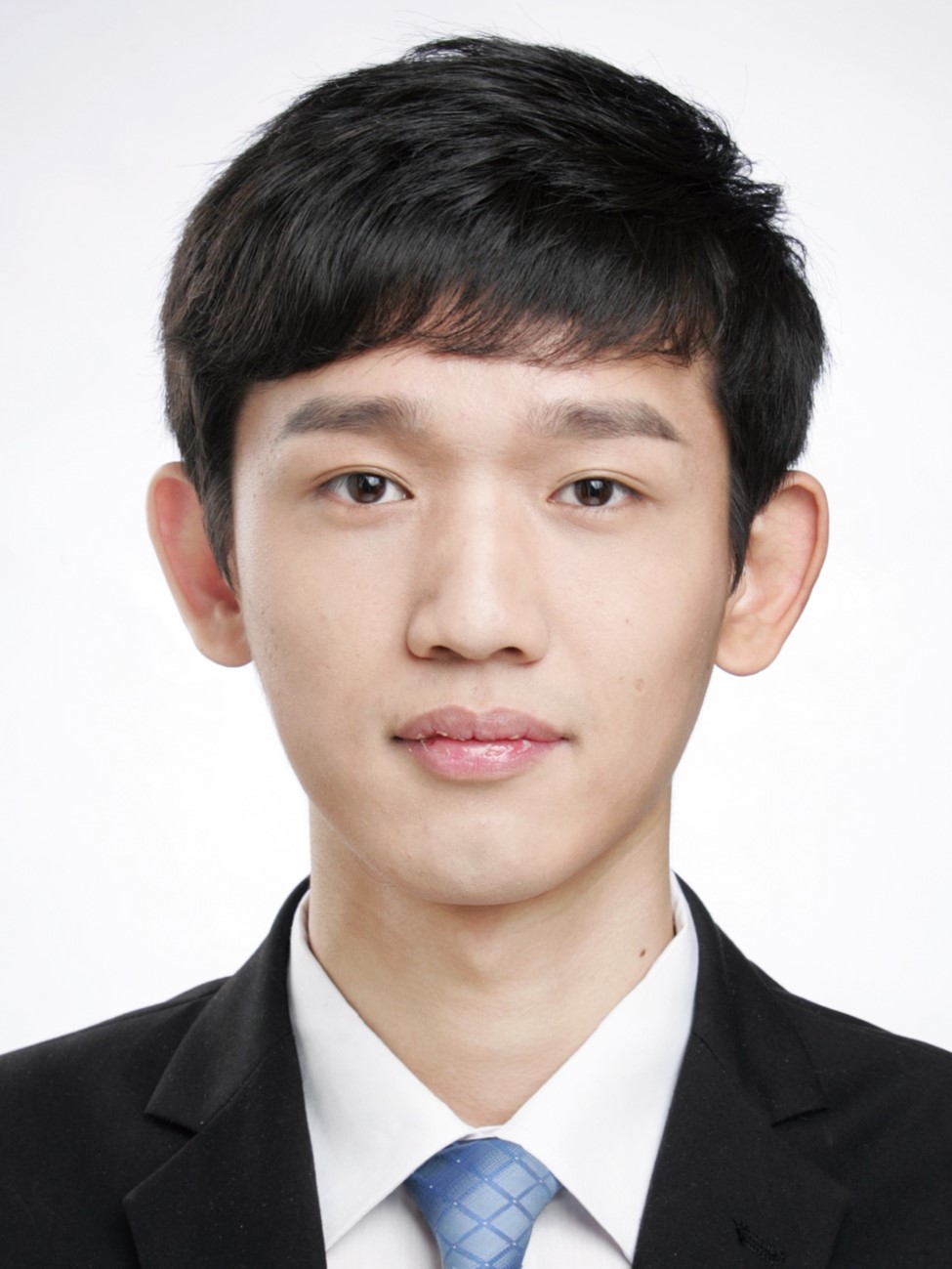}}]{Yuecong Min}
received the BS degree in information and computing science from Shandong University in 2017. He received the PhD degree in computer science from the Institute of Computing Technology, Chinese Academy of Sciences in 2024. He is currently a postdoctoral researcher with the Institute of Computing Technology, Chinese Academy of Sciences. His research interests are in computer vision, pattern recognition, and machine learning. He especially focuses on sign language processing, vision-language modeling and the related research topics.
\end{IEEEbiography}

\begin{IEEEbiography}[{\includegraphics[width=1in,height=1.25in,clip,keepaspectratio]{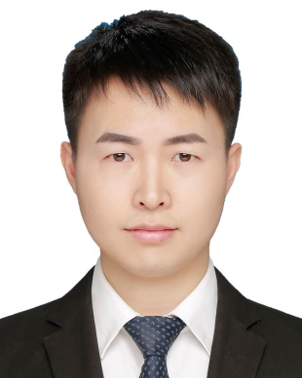}}]{Jie Zhang}
received the Ph.D. degree from the University of Chinese Academy of Sciences (CAS), Beijing, China. He is currently an Associate Professor with the Institute of Computing Technology, CAS. His research interests include computer vision, pattern recognition, machine learning, particularly include adversarial attacks and defenses,  domain generalization, AI safety and trustworthiness.
\end{IEEEbiography}

\begin{IEEEbiography}[{\includegraphics[width=1in,height=1.25in,clip,keepaspectratio]{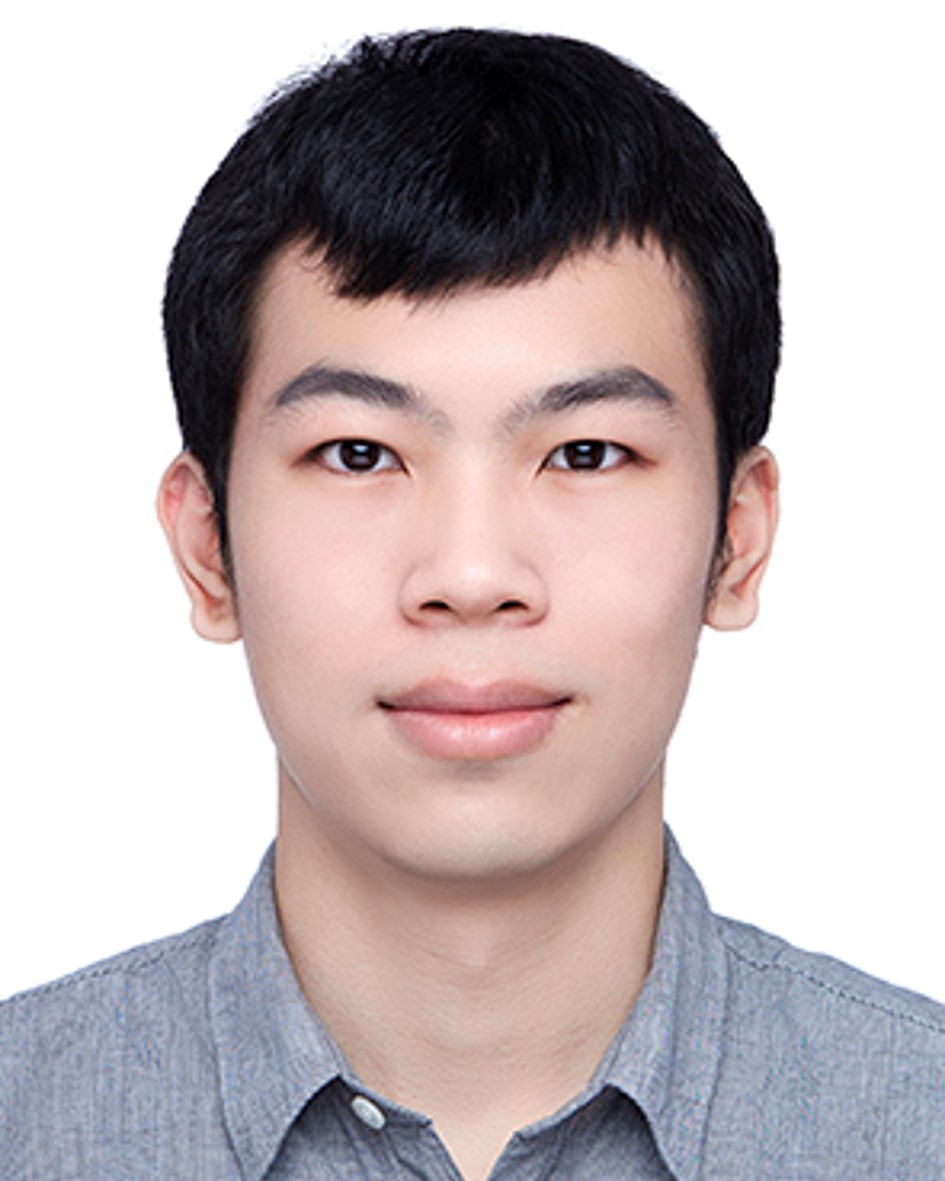}}]{Zheng Yuan}
received the B.S. degree from University of Chinese Academy of Sciences in 2019. He is currently pursuing the Ph.D. degree from University of Chinese Academy of Sciences. His research interest includes adversarial example and model robustness. He has authored several academic papers in international conferences including ICCV/ECCV/ICPR and journals including TPAMI/IJCV.
\end{IEEEbiography}

\begin{IEEEbiography}[{\includegraphics[width=1in,height=1.25in,clip,keepaspectratio]{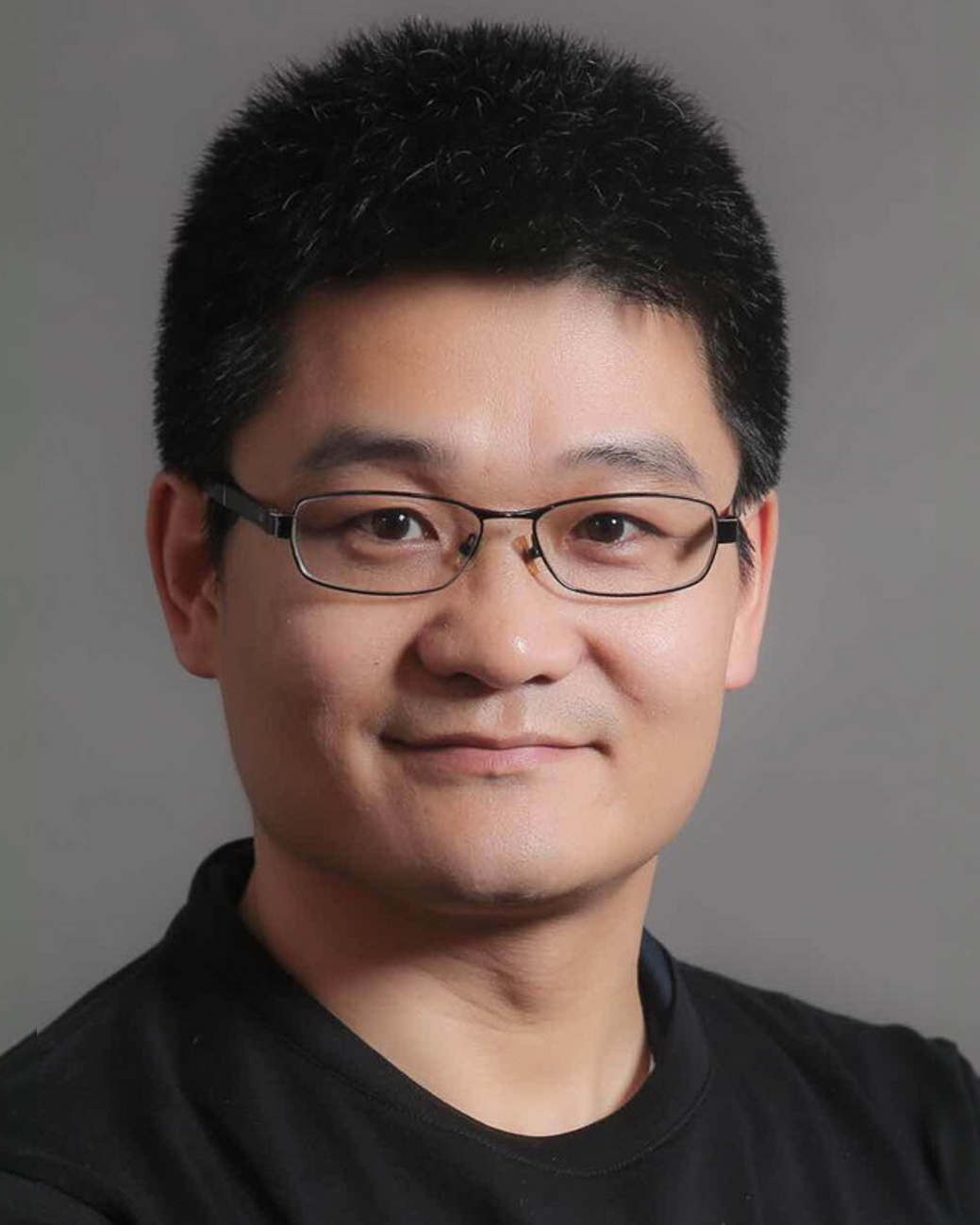}}]{Shiguang Shan}
(Fellow, IEEE) received the Ph.D. degree in computer science from the Institute of Computing Technology (ICT), Chinese Academy of Sciences (CAS), Beijing, China, in 2004. He has been a Full Professor with ICT since 2010, where he is currently the Director of the Key Laboratory of Intelligent Information Processing, CAS. His research interests include signal processing, computer vision, pattern recognition, and machine learning. He has published more than 300 articles in related areas. He served as the General Co-Chair for IEEE Face and Gesture Recognition 2023, the General Co-Chair for Asian Conference on Computer Vision (ACCV) 2022, and the Area Chair of many international conferences, including CVPR, ICCV, AAAI, IJCAI, ACCV, ICPR, and FG. He was/is an Associate Editor of several journals, including IEEE Transactions on Image Processing, Neurocomputing, CVIU, and PRL. He was a recipient of the China's State Natural Science Award in 2015 and the China’s State S\&T Progress Award in 2005 for his research work.
\end{IEEEbiography}

\begin{IEEEbiography}[{\includegraphics[width=1in,height=1.25in,clip,keepaspectratio]{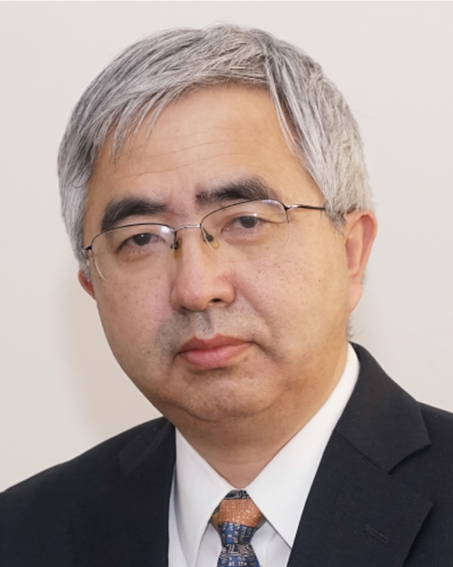}}]{Xilin Chen} (Fellow, IEEE) is currently a Professor with the Institute of Computing Technology, Chinese
 Academy of Sciences (CAS). He has authored one
 book and more than 400 articles in refereed journals
 and proceedings in the areas of computer vision,
 pattern recognition, image processing, and multi
modal interfaces. He is a fellow of the ACM,
 IAPR, and CCF. He is also an Information Sciences
 Editorial Board Member of Fundamental Research,
 an Editorial Board Member of Research, a Senior
 Editor of the Journal of Visual Communication and
 Image Representation, and an Associate Editor-in-Chief of the Chinese Jour
nal of Computers and Chinese Journal of Pattern Recognition and Artificial
 Intelligence. He served as an organizing committee member for multiple
 conferences, including the General Co-Chair of FG 2013/FG 2018, VCIP
 2022, the Program Co-Chair of ICMI 2010/FG 2024, and an Area Chair of
 ICCV/CVPR/ECCV/NeurIPS for more than ten times.
\end{IEEEbiography}

\clearpage

\section*{Supplementary Materials}

These supplementary materials provide additional details, results, and analyses to support the main paper, including (A) implementation details, (B) complete results of adapted T2I attacks, (C) comprehensive text editing results, (D) impact of model hyperparameters, (E) in-depth analysis of temporal attacks, (F) construction details of T2VAttackBench, and (G) additional visualization results.

\begin{description}
    \item[{\hyperref[sup:details]{ \textcolor{blue}{\underline{A}} }}]  We provide the implementation details, including victim model configurations and attack hyperparameter settings.

    \item[{\hyperref[sup:qf]{ \textcolor{blue}{\underline{B}} }}]       We report the complete results of QueryFreeAdv adapted to four T2V models across three optimization variants, including its stealthiness metrics, and analyze the decoupling between encoder-space displacement and video-level degradation.

    \item[{\hyperref[sup:edit]{ \textcolor{blue}{\underline{C}} }}]     We expand position-wise studies (first, middle, last, random, important) across four text editing operations.
    
    \item[{\hyperref[sup:impact]{ \textcolor{blue}{\underline{D}} }}]   We investigate the influence of key model hyperparameters on semantic and temporal objectives to validate our standardized evaluation protocol.
    
    \item[{\hyperref[sup:pure]{ \textcolor{blue}{\underline{E}} }}]     We provide an in-depth analysis of temporal attacks, focusing on degrading temporal dynamics while preserving semantic content.

    \item[{\hyperref[sup:dataset]{ \textcolor{blue}{\underline{F}} }}]  We provide the generation prompts for our T2VAttackBench dataset, along with release information.

    \item[{\hyperref[sup:vis]{ \textcolor{blue}{\underline{G}} }}]      We present additional qualitative results on HunyuanVideo.
\end{description}

\subsection{Implementation Details}
\label{sup:details}

Among the victim models, ModelScope~\cite{wang2023modelscope}, CogVideoX~\cite{yang2024cogvideox}, and HunyuanVideo~\cite{kong2024hunyuanvideo} are implemented using the Diffusers library~\cite{von-platen-etal-2022-diffusers}, while Open-Sora~\cite{zheng2024open} is based on its official implementation. Specifically, we utilize the \textit{text-to-video-ms-1.7b} version for ModelScope, the \textit{CogVideoX-2b} version for CogVideoX, the \textit{Open-Sora v1.2} version for Open-Sora, and the latest \textit{HunyuanVideo-13B} for HunyuanVideo.
Due to GPU memory limitations, we employ model parallelism for parallel inference with Open-Sora on two GPUs. For HunyuanVideo, we utilize a combination of unified sequence parallelism and FP8 quantization for parallel inference on two GPUs. All videos from the four victim models are generated on NVIDIA RTX 3090 GPUs. Their detailed inference configurations are provided in Table~\ref{tab:victims}.

For our attack methods, the following settings are used. In T2VAttack-S, $X_{\setminus x_i}$ (the prompt $X$ removing $x_i$) is computed by replacing the target word $x_i$ with the \textit{\texttt{<UNK>}} token. The calculation of the candidate synonyms' similarity is performed using a 6-token window. The synonym filtering threshold $\theta$ is set to 0.8; for prompts shorter than six tokens, both the window length and the threshold $\theta$ are halved. The objective score thresholds $\tau_s$ and $\tau_t$ for the semantic and temporal objectives are set to 0.5 and 0.1, respectively. The maximum word substitution limit is set to 3. In T2VAttack-I, we employ only a first-level prefix with 64 queries. 
The word vocabulary is derived from GloVe~\cite{pennington2014glove}, filtered by SpellChecker libraries, and contains 74,495 valid words.
Regarding the calculation of attack scores, video-text similarity scores are scaled to the range [0, 1] based on the reference interval [0, 0.364] following VBench~\cite{huang2024vbench}, and optical flow scores are clamped to the range [0, 10] and then normalized to [0, 1]. To compute formal similarity, the Levenshtein edit distance is first calculated using dynamic programming, followed by 1 minus the normalized Levenshtein distance; semantic similarity is calculated based on the embeddings from the \textit{all-MiniLM-L6-v2} model from SentenceTransformer.

\begin{table}[t]
    \centering
    \caption{Specifications of T2V victim models employed in our experiments.
    The table summarizes key hyperparameters of four models (ModelScope, CogVideoX, Open-Sora, and HunyuanVideo), including resolution, frames per second (FPS), total frame count, video length, and inference steps, which are crucial for understanding model behavior under adversarial attack conditions.
    }
    \renewcommand{\arraystretch}{1.2} 
    \begin{tabular}{lcccccc} 
        \toprule
        \textbf{Victim Models} & 
        \textbf{Resolution} & 
        \textbf{FPS} & 
        \textbf{\begin{tabular}[c]{@{}c@{}}Frame\\ Count\end{tabular}} & 
        \textbf{\begin{tabular}[c]{@{}c@{}}Video\\ Length\end{tabular}} & 
        \textbf{\begin{tabular}[c]{@{}c@{}}Inference\\ Steps\end{tabular}} \\
        \midrule
        ModelScope   & $720\!\times\!480$ & 8  & 16 & 2.0 s & 25 \\
        CogVideoX    & $256\!\times\!256$ & 8  & 16 & 2.0 s & 25 \\
        Open-Sora    & $426\!\times\!240$ & 24 & 49 & 2.0 s & 25 \\
        HunyuanVideo & $512\!\times\!320$ & 24 & 49 & 2.0 s & 25 \\
        \bottomrule
    \end{tabular}
    \label{tab:victims}
\end{table}

\begin{table*}[!h]
    \centering
    \caption{Complete results of QueryFreeAdv~\cite{zhuang2023pilot} adapted to four T2V models under the semantic and temporal objectives. We report the efficacy and stealthiness metrics across its three optimization variants (Greedy, Genetic, and PGD).
    }
    \renewcommand{\arraystretch}{1.1}
    \label{tab:qf_full}
    \begin{tabular}{cc|cc|ccc}
        \toprule
        \textbf{\makecell{Victim\\ Models}} & 
        \textbf{\makecell{Optimization\\ Variant}} & 
        \textbf{\makecell{Semantic\\ Sim. $\uparrow$}} & 
        \textbf{\makecell{Formal\\ Sim. $\uparrow$}} & 
        \textbf{\makecell{Original\\ Score}} & 
        \textbf{\makecell{Adversarial\\ Score}} & 
        \textbf{Difference $\uparrow$} \\
        \midrule
        \rowcolor{gray!20}
        \multicolumn{7}{c}{\textbf{Semantic Attack}} \\
        \midrule
        \multirow{3}{*}{ModelScope}
        & Greedy   & 0.94 & 0.83 & 81.1 & 77.8 & 3.3 \\
        & Genetic   & 0.93 & 0.85 & 81.1 & 78.5 & 2.6 \\
        & PGD   & 0.93 & 0.85 & 81.1 & 78.8 & 2.3 \\
        \midrule
        \multirow{3}{*}{CogVideoX}
        & Greedy    & 0.92 & 0.81 & 79.8 & 77.0 & 2.8 \\
        & Genetic    & 0.93 & 0.81 & 79.8 & 77.9 & 1.9 \\
        & PGD    & 0.94 & 0.81 & 79.8 & 77.4 & 2.4 \\
        \midrule
        \multirow{3}{*}{Open-Sora}
        & Greedy    & 0.93 & 0.82 & 80.1 & 77.7 & 2.4 \\
        & Genetic    & 0.92 & 0.81 & 80.1 & 79.7 & 0.4 \\
        & PGD    & 0.93 & 0.80 & 80.1 & 78.0 & 2.1 \\
        \midrule
        \multirow{3}{*}{HunyuanVideo}
        & Greedy & 0.92 & 0.85 & 83.0 & 82.3 & 0.7 \\
        & Genetic & 0.92 & 0.82 & 83.0 & 82.4 & 0.6 \\
        & PGD & 0.92 & 0.82 & 83.0 & 82.2 & 0.8 \\
        \midrule
        \midrule
        \rowcolor{gray!20}
        \multicolumn{7}{c}{\textbf{Temporal Attack}} \\
        \midrule
        \multirow{3}{*}{ModelScope}
        & Greedy   & 0.88 & 0.75 & 45.2 & 45.2 & 0.0 \\
        & Genetic   & 0.88 & 0.77 & 45.2 & 44.0 & 1.2 \\
        & PGD   & 0.86 & 0.80 & 45.2 & 39.8 & 5.4 \\
        \midrule
        \multirow{3}{*}{CogVideoX}
        & Greedy    & 0.87 & 0.70 & 83.6 & 78.5 & 5.1 \\
        & Genetic    & 0.87 & 0.70 & 83.6 & 78.0 & 5.6 \\
        & PGD    & 0.88 & 0.71 & 83.6 & 76.9 & 6.7 \\
        \midrule
        \multirow{3}{*}{Open-Sora}
        & Greedy    & 0.87 & 0.71 & 67.9 & 72.2 & $-$4.3 \\
        & Genetic    & 0.86 & 0.70 & 67.9 & 70.1 & $-$2.2 \\
        & PGD    & 0.88 & 0.69 & 67.9 & 69.3 & $-$1.4 \\
        \midrule
        \multirow{3}{*}{HunyuanVideo}
        & Greedy & 0.86 & 0.74 & 72.1 & 76.3 & $-$4.2 \\
        & Genetic & 0.87 & 0.70 & 72.1 & 73.5 & $-$1.4 \\
        & PGD & 0.86 & 0.71 & 72.1 & 74.5 & $-$2.4 \\
        \bottomrule
    \end{tabular}
    \vspace{-8pt}
\end{table*}

\begin{figure}[b]
    \centering
    \includegraphics[width=0.92\linewidth]{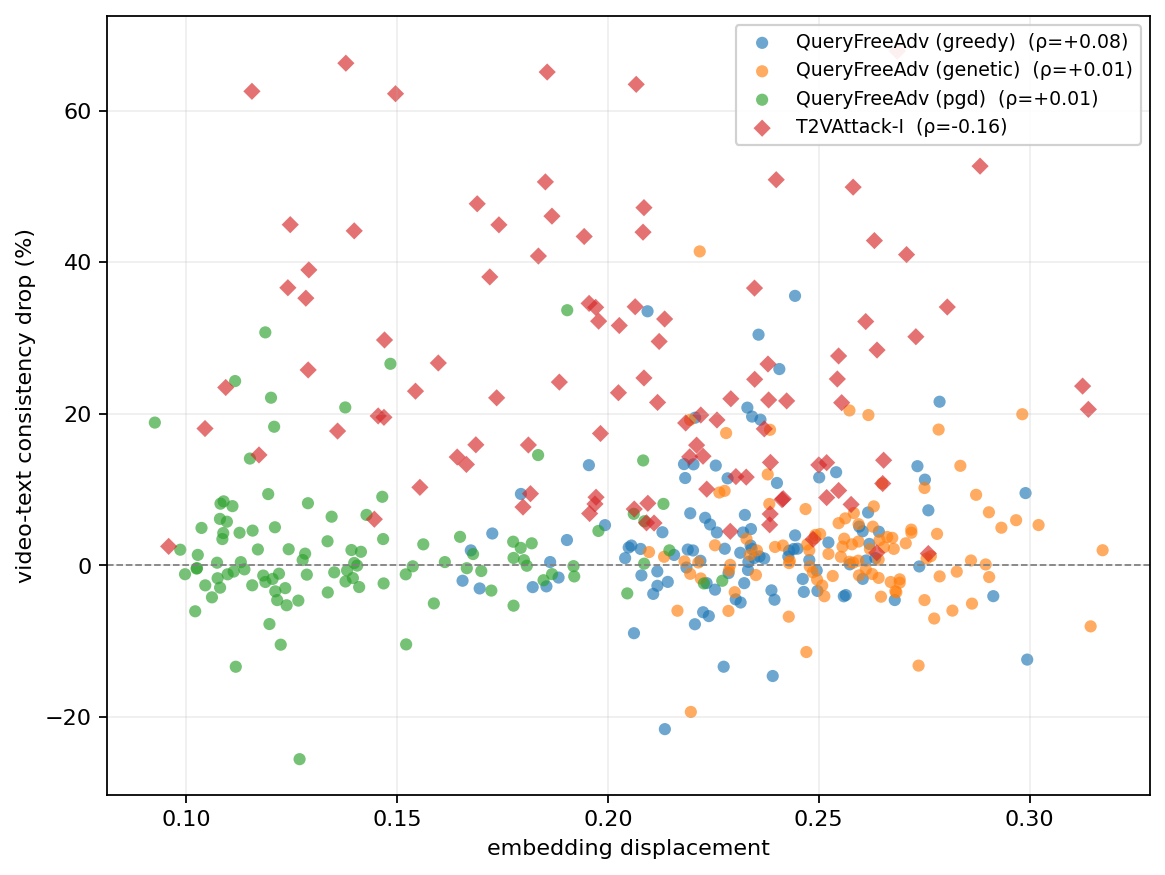}
    \caption{Correlation between the QueryFreeAdv objective (prompt embedding displacement) and the true video objective (video-text consistency drop) on ModelScope. Although all three QueryFreeAdv variants successfully induce text-embedding displacements, these geometric shifts exhibit a near-zero rank correlation with the actual video-level drop. In contrast, T2VAttack-I achieves substantially larger drops at comparable displacements.
    }
    \label{fig:displacement_drop}
\end{figure}

\subsection{Complete Results of Adapted T2I Attacks}
\label{sup:qf}
Table~\ref{tab:qf_full} reports the complete results of QueryFreeAdv~\cite{zhuang2023pilot} adapted to the four T2V victim models, complementing the best-variant summary in the main paper. For each victim model, we attack its specific text encoder (CLIP for ModelScope, T5 for CogVideoX and Open-Sora, and Llama for HunyuanVideo) using the official five-character perturbation protocol across three optimization variants: greedy search, genetic algorithm, and PGD~\cite{madry2018towards}.
In terms of attack efficacy, the adapted T2I attacks exert negligible impact on T2V generation. Across all variants and models, the semantic-score reductions do not exceed 3.3 points. Temporally, the attacks fail to suppress motion and often inadvertently increase the dynamic degree of the generated videos, confirming that encoder-level perturbations provide no reliable control over spatiotemporal behavior. Regarding stealthiness, the semantic similarity remains high (0.86--0.94) as QueryFreeAdv preserves existing words. However, the formal similarity drops comparatively lower (0.69--0.85) because the fragmentation of the appended random characters during tokenization degrades token-level formal similarity.

To further explain this inefficacy, Fig.~\ref{fig:displacement_drop} contrasts the objective optimized by QueryFreeAdv (the cosine displacement of the prompt embedding) with the true video objective of interest (the drop in video-text consistency) on ModelScope. 
Although all three QueryFreeAdv variants successfully induce significant text-embedding displacements (yielding mean cosine displacement of 0.23, 0.26, and 0.14 for the greedy, genetic, and PGD variants, respectively) comparable to our T2VAttack-I (0.21), their resulting video-level degradation remains negligible. 
Crucially, these geometric displacements exhibit a near-zero rank correlation with the actual video-text consistency drop within each variant (Spearman's $\rho=+0.08$, $+0.01$, and $+0.01$ for greedy, genetic, and PGD, respectively; all $p>0.4$). 
In stark contrast, T2VAttack-I, which is directly optimized via the video-level metric, achieves a substantially larger mean drop (25.1 versus 2.3--3.3 points for QueryFreeAdv) at a comparable displacement. 
This inherent decoupling precisely explains the failure of adapted T2I attacks and validates the necessity of the video-level optimization objectives adopted in T2VAttack.

\subsection{Comprehensive Text Editing Operations Results}
\label{sup:edit}

Table~\ref{table:attack_performance_by_position_merged} extends the analysis of text editing operations on ModelScope in Table~\ref{tab:attackers_position} by evaluating additional edit positions: middle, last, and random, in addition to first and important positions. Consistent with the findings in the main paper, insertion operations remain highly effective across all positions, with the strongest influence at the first position. Moreover, substitution and deletion operations have a stronger impact when applied to the important position, whereas these edits at the first, middle, last, and random positions yield limited degradation. Furthermore, reordering operations continue to exhibit marginal effects, including the newly considered middle, last, and random, confirming its limited adversarial impact.
Overall, these extended experiments further corroborate the main text analysis that insertion operations are broadly effective across different positions, while substitution and deletion operations rely heavily on targeting critical words.

\begin{table*}[t]
    \centering
    \caption{
    Comparison of various text editing operations across different positions. The table specifically introduces additional edit positions, first, middle, last, random, and important, to provide more comprehensive results.
    }
    \renewcommand{\arraystretch}{1.1}
    \begin{tabular}{ccccccc}
        \toprule
        \textbf{\begin{tabular}[c]{@{}c@{}}Text\\ Editing\end{tabular}} & 
        \textbf{Position} & 
        \textbf{\begin{tabular}[c]{@{}c@{}}Semantic\\ Similarity $\uparrow$\end{tabular}} & 
        \textbf{\begin{tabular}[c]{@{}c@{}}Formal\\ Similarity $\uparrow$\end{tabular}} & 
        \textbf{\begin{tabular}[c]{@{}c@{}}Original\\ Score\end{tabular}} & 
        \textbf{\begin{tabular}[c]{@{}c@{}}Adversarial\\ Score\end{tabular}} & 
        \textbf{\begin{tabular}[c]{@{}c@{}}Semantic\\ Difference $\uparrow$\end{tabular}} \\
        \midrule
        \multirow{5}{*}{Insertion}
        & first & 0.89 & 0.91 & 81.1 & 56.0 & 25.1 \\
        & middle & 0.92 & 0.91 & 81.1 & 59.5 & 21.6 \\
        & last & 0.93 & 0.91 & 81.1 & 61.1 & 20.0 \\
        & random & 0.91 & 0.91 & 81.1 & 60.0 & 21.1 \\
        & important & 0.93 & 0.91 & 81.1 & 57.8 & 23.3 \\
        \cline{1-7}
        \multirow{5}{*}{Substitution}
        & first & 0.89 & 0.91 & 81.1 & 77.1 & 4.0 \\
        & middle & 0.91 & 0.91 & 81.1 & 74.0 & 7.1 \\
        & last & 0.89 & 0.91 & 81.1 & 73.7 & 7.4 \\
        & random & 0.86 & 0.91 & 81.1 & 71.1 & 10.0 \\
        & important & 0.82 & 0.91 & 81.1 & 59.5 & 21.6 \\
        \cline{1-7}
        \multirow{5}{*}{Deletion}
        & first & 0.97 & 0.91 & 81.1 & 78.7 & 2.4 \\
        & middle & 0.96 & 0.91 & 81.1 & 76.6 & 4.5 \\
        & last & 0.95 & 0.91 & 81.1 & 74.9 & 6.2 \\
        & random & 0.93 & 0.91 & 81.1 & 73.8 & 7.3 \\
        & important & 0.89 & 0.91 & 81.1 & 59.3 & 21.8 \\
        \cline{1-7}
        \multirow{5}{*}{Reordering}
        & first & 0.97 & 0.82 & 81.1 & 77.8 & 3.3 \\
        & middle & 0.97 & 0.82 & 81.1 & 76.1 & 5.0 \\
        & last & 0.98 & 0.82 & 81.1 & 73.9 & 7.2 \\
        & random & 0.98 & 0.82 & 81.1 & 76.3 & 4.8 \\
        & important & 0.98 & 0.82 & 81.1 & 73.2 & 7.9 \\
        \bottomrule
    \end{tabular}
    \label{table:attack_performance_by_position_merged}
\end{table*}

\subsection{Impact of Model Hyperparameters}
\label{sup:impact}

Existing T2V models often exhibit different hyperparameters, and more recent works typically generate higher-quality videos characterized by increased resolution and longer duration. 
To investigate the influence of model hyperparameters on our attack evaluation metrics, we focus on three critical hyperparameters in the video generation process: resolution, frame count, and inference steps. The experimental results in Table~\ref{tab:combined_ablation} show the impact of these model hyperparameters on semantic and temporal objectives for HunyuanVideo.
Surprisingly, both the semantic and temporal objectives demonstrate robustness to changes in model hyperparameters, with the exception of the influence of video frame count on the temporal objective. Therefore, for consistency, we generate 2.0-second videos across all models, adhering to the model's inherent frame rate setting. More detailed inference configurations are provided in Table~\ref{tab:victims}.

\begin{table*}[htbp]
    \centering
    \caption{Impact of model hyperparameters on semantic and temporal objectives: (a) input resolution, (b) frame count, (c) inference steps.
    }
    \vspace{-8pt}
    \label{tab:combined_ablation}
    \subfloat{
        \begin{minipage}[t]{0.32\textwidth}
            \centering
            {\renewcommand{\arraystretch}{1.2}
            {\small (a) Resolution}
            \par\vspace{5pt}
            \resizebox{\linewidth}{!}{
            \begin{tabular}{ccc}
                \toprule
                \textbf{\makecell{Input\\ Resolution}} & \textbf{\makecell{Semantic\\ Objective}} & \textbf{\makecell{Temporal\\ Objective}} \\
                \midrule
                320$\times$240 & 81.44 & 77.79 \\
                512$\times$320 & 83.00 & 72.19 \\
                720$\times$480 & 82.85 & 74.25 \\
                \bottomrule
            \end{tabular}
            }
            }
        \end{minipage}
    }
    \hfill
    \subfloat{
        \begin{minipage}[t]{0.28\textwidth}
            \centering
            {
            {\small (b) Frame Count}
            \par\vspace{5pt}
            \resizebox{\linewidth}{!}{
            \begin{tabular}{ccc}
                \toprule
                \textbf{\makecell{Frame\\ Count}} & \textbf{\makecell{Semantic\\ Objective}} & \textbf{\makecell{Temporal\\ Objective}} \\
                \midrule
                17 & 80.45 & 35.08 \\
                25 & 82.11 & 49.32 \\
                49 & 83.00 & 72.19 \\
                97 & 82.26 & 72.46 \\
                \bottomrule
            \end{tabular}
            }
            }
        \end{minipage}
    }
    \hfill
    \subfloat{
        \begin{minipage}[t]{0.32\textwidth}
            \centering
            {\renewcommand{\arraystretch}{1.2}
            {\small (c) Inference Steps}
            \par\vspace{5pt}
            \resizebox{\linewidth}{!}{
            \begin{tabular}{ccc}
                \toprule
                \textbf{\makecell{Inference\\ Steps}} & \textbf{\makecell{Semantic\\ Objective}} & \textbf{\makecell{Temporal\\ Objective}} \\
                \midrule
                13 & 81.70 & 68.40 \\
                25 & 83.00 & 72.19 \\
                50 & 82.62 & 71.09 \\
                \bottomrule
            \end{tabular}
            }
            }
        \end{minipage}
    }
\end{table*}

\subsection{In-depth Analysis of Temporal Attacks}
\label{sup:pure}

When assessing the temporal attack, we observe that reducing the temporal dynamics of videos may also inadvertently disrupt their semantic information. To address this issue, we conduct a preliminary exploration into the temporal attack, which aims to degrade the temporal dynamics of the generated video while preserving its semantic content. Specifically, we introduce an additional semantic constraint into our temporal objective, ensuring that the adversarial video remains semantically similar to its original static video equivalent. 
To achieve this, we first generate an original video $G(X)$ based on the original prompt, and then create a static video $SG(X)$ by removing temporal displacements, which can be simplified by stacking the middle frame of the original video $G(X)$. The new temporal objective $\hat{f}_t$ is designed to reduce temporal dynamics while preserving semantic information:
\setcounter{equation}{5}
\begin{equation}
\hat{f}_t = \sum_{\tau=1}^{T-1} \rho(G(X')_\tau, G(X')_{\tau+1}) - \lambda \, sim(E_v(SG(X)), E_v(G(X'))),
\end{equation}
where $SG(\cdot)$ refers to the static video generation process and $\lambda$ is a hyperparameter to balance the two components.
This constrained objective allows for an effective attack on temporal dynamics without significantly altering the semantic integrity of the generated video, thereby enabling a precise evaluation of temporal adversarial robustness.

\clearpage
Table~\ref{tab:attack_metric_lambda} shows the impact of the hyperparameter $\lambda$ on both attack effectiveness and semantic preservation, based on temporal attacks on ModelScope. The Video-Video Similarity metric measures the semantic similarity between the adversarial video and the original video. Before the attack, the adversarial prompt causes no perturbations, meaning the Video-Video Similarity is at its maximum value of 100. The results show that as the hyperparameter $\lambda$ increases, the Video-Video Similarity also increases, while the temporal dynamics metric decreases. This suggests that preserving more semantic information hinders the attack on temporal dynamics. 

\begin{table*}[t]
    \centering
    \caption{Ablation studies of the hyperparameter $\lambda$ on temporal attack, showing the trade-off between maintaining semantic fidelity and degrading temporal dynamics. 
    }
    \renewcommand{\arraystretch}{1.1}
    \begin{tabular}{c|cc|ccc|ccc} 
        \toprule
        \multirow{3}{*}{\textbf{\begin{tabular}[c]{@{}c@{}}Hyperparameter $\lambda$ \end{tabular}}} & 
        \multirow{3}{*}{\textbf{\begin{tabular}[c]{@{}c@{}}Semantic\\ Similarity $\uparrow$\end{tabular}}} & 
        \multirow{3}{*}{\textbf{\begin{tabular}[c]{@{}c@{}}Formal\\ Similarity $\uparrow$\end{tabular}}} & 
        \multicolumn{3}{c}{\textbf{Temporal Metric}} & 
        \multicolumn{3}{c}{\textbf{Video-Video Similarity}} \\ 
        \cmidrule(lr){4-6} \cmidrule(lr){7-9} 
        & & & \multirow{2}{*}{\textbf{\begin{tabular}[c]{@{}c@{}}Original\\ Score\end{tabular}}} & 
        \multirow{2}{*}{\textbf{\begin{tabular}[c]{@{}c@{}}Adversarial\\ Score\end{tabular}}} & 
        \multirow{2}{*}{\textbf{\makecell{Temporal\\ Difference $\uparrow$}}} & 
        \multirow{2}{*}{\textbf{\begin{tabular}[c]{@{}c@{}}Original\\ Score\end{tabular}}} & 
        \multirow{2}{*}{\textbf{\begin{tabular}[c]{@{}c@{}}Adversarial\\ Score\end{tabular}}} & 
        \multirow{2}{*}{\textbf{\makecell{Video-Video \\ Difference $\downarrow$}}} \\ 
        & & & & & & & & \\
        \midrule
        0 & 0.83 & 0.86 & 45.2 & 3.4 & 41.8 & 100.0 & 67.0 & 33.0 \\
        0.1 & 0.84 & 0.86 & 45.2 & 3.6 & 41.6 & 100.0 & 76.9 & 23.1 \\
        0.2 & 0.84 & 0.86 & 45.2 & 3.9 & 41.3 & 100.0 & 79.3 & 20.7 \\
        0.4 & 0.84 & 0.86 & 45.2 & 5.2 & 40.0 & 100.0 & 83.5 & 16.5 \\
        0.8 & 0.84 & 0.86 & 45.2 & 6.8 & 38.4 & 100.0 & 86.2 & 13.8 \\
        \bottomrule
    \end{tabular}
    \label{tab:attack_metric_lambda}
\end{table*}

\subsection{Construction Details of T2VAttackBench}
\label{sup:dataset}

Although VBench~\cite{huang2024vbench} provides a comprehensive benchmark for evaluating video generative models, its results reveal that almost all state-of-the-art models still fail to produce sufficiently complex and coherent videos. This limitation makes the direct use of VBench suboptimal for adversarial evaluation, as inherent performance deficits may obscure the true impact of attacks. To address this, we construct \textbf{T2VAttackBench}. Unlike conventional benchmarks that gauge only average-case generative quality, T2VAttackBench is engineered to expose worst-case vulnerabilities along the two dimensions that jointly characterize video quality: semantic alignment and temporal dynamics. Accordingly, it encompasses two subsets, a \textit{Semantic} subset of 105 prompts and a \textit{Temporal} subset of 52 prompts, both distilled from VBench through GPT-4o refinement and retained only after a stringent multi-model consensus-filtering protocol (Sect.~\ref{sec:exp}). This curation ensures that, under benign conditions, every prompt elicits videos with strong semantic consistency and pronounced motion, thereby furnishing a clean substrate that isolates attack-induced degradation from a model's intrinsic limitations. To facilitate future research on T2V robustness, the complete benchmark, together with the generated videos and evaluation scripts, is publicly available at \href{https://drive.google.com/drive/folders/1bjRIgUto0NzTUG52KZIt706PnjQPDNtM?usp=sharing}{\textcolor{blue}{\underline{this repository}}}.

\subsubsection{Semantic Dataset}
The Semantic subset encompasses visually rich, compositional, and frequently counterfactual scenes whose faithful synthesis hinges on precise cross-modal grounding, rendering them sensitive probes of a model's semantic fidelity.

Prompt Template:

\begin{quote}
"As an expert linguist, your task is to create a high-quality textual dataset with \textbf{rich visual content} for text-to-video generation tasks. Following the style and structure demonstrated in the reference examples provided below, generate 200 textual prompts. Ensure each prompt has low perplexity, clear semantics, and high comprehensibility, thus enabling \textbf{the generation of visually engaging and rich video content}. The prompts should cover diverse categories and scenarios, including but not limited to animals, architecture, food, humans, lifestyles, plants, scenery, and vehicles.
  
Reference samples:

Close up of grapes on a rotating table.

Turtle swimming in ocean.

A storm trooper vacuuming the beach.

\ldots
"
\end{quote}

Its 105 constituent prompts are:
{\small
\begin{enumerate}
\item close up of grapes on a rotating table
\item a storm trooper vacuuming the beach
\item a panda standing on a surfboard in the ocean in sunset
\item two pandas discussing an academic paper
\item a fat rabbit wearing a purple robe walking through a fantasy landscape
\item a koala bear playing piano in the forest
\item an astronaut flying in space
\item a bigfoot walking in the snowstorm
\item a squirrel eating a burger
\item snow rocky mountains peaks canyon. snow blanketed rocky mountains surround and shadow deep canyons. the canyons twist and bend through the high elevated mountain peaks
\item a teddy bear is swimming in the ocean
\item happy dog wearing a yellow turtleneck, studio, portrait, facing camera, dark background
\item campfire at night in a snowy forest with starry sky in the background
\item a 3D model of a 1800s victorian house
\item robot dancing in Times Square
\item busy freeway at night
\item an astronaut is riding a horse in the space in a photorealistic style
\item vampire makeup face of beautiful girl, red contact lenses
\item a teddy bear is playing drum kit in NYC Times Square
\item a corgi is playing drum kit
\item a raccoon is playing the electronic guitar
\item a jellyfish floating through the ocean, with bioluminescent tentacles
\item a Mars rover moving on Mars
\item a panda drinking coffee in a cafe in Paris
\item a space shuttle launching into orbit, with flames and smoke billowing out from the engines
\item a steam train moving on a mountainside
\item cinematic shot of Van Gogh's selfie, Van Gogh style
\item iron Man flying in the sky
\item a boat sailing leisurely along the Seine River with the Eiffel Tower in background
\item a cat eating food out of a bowl
\item a cat wearing sunglasses at a pool
\item a cute fluffy panda eating Chinese food in a restaurant
\item a cute raccoon playing guitar in a boat on the ocean
\item a happy fuzzy panda playing guitar nearby a campfire, snow mountain in the background
\item a panda cooking in the kitchen
\item a polar bear is playing guitar
\item a raccoon dressed in suit playing the trumpet, stage background
\item an epic tornado attacking above a glowing city at night, the tornado is made of smoke
\item an oil painting of a couple in formal evening wear going home get caught in a heavy downpour with umbrellas
\item clown fish swimming through the coral reef
\item a flamingo balancing on one leg in a pink-hued lagoon at dawn, reflections rippling softly
\item a jellyfish pulsating with bioluminescent lights in the deep ocean abyss
\item aurora borealis shimmering over a frozen lake surrounded by snow-dusted pines
\item a volcanic eruption spewing lava into the ocean, steam clouds rising dramatically
\item a hot air balloon drifting over patchwork fields at sunrise, aerial view
\item a starry night sky rendered in Van Gogh's brushstrokes, dynamic swirling motion
\item a cybernetic tiger prowling through a glitch-art digital landscape
\item a Renaissance-style oil painting of a robot holding a rose, textures vivid
\item a ballet dancer twirling in a snow globe, glitter swirling in golden light
\item a child flying a dragon-shaped kite in a windy autumn park, leaves swirling
\item a snow globe containing a miniature NYC during a blizzard, close-up
\item a neon ``Open 24/7'' sign flickering in a rainy alley, puddle reflections
\item a kangaroo boxing a robot in a futuristic arena, crowd cheering
\item a birthday cake with candle-fireworks igniting when blown
\item a potter shaping clay on a wheel, hands muddy, close-up
\item a knight riding through sunflowers, armor gleaming in sunlight
\item a DJ controlling lightning bolts with turntables in a stormy sky
\item a yoga instructor meditating on a floating rock over a misty valley
\item a redwood tree doorway leading to a hidden fairy village, fireflies
\item a savanna sunset with elephant silhouettes against acacia trees
\item a meteor shower over a desert, cacti outlined by streaks of light
\item a horse-drawn carriage in a snow-covered village, lanterns glowing
\item a unicorn galloping on clouds, rainbows forming in its wake
\item a steampunk airship battling mechanical dragons over factories
\item a minimalist desert with one cactus casting a long shadow
\item an hourglass with star-sand flowing in zero gravity
\item a majestic eagle soaring over a rugged mountain range at sunset, with golden light illuminating its wings
\item a lion roaring atop a savannah hill at dawn, with the first rays of light breaking over the horizon
\item a squirrel munching on a tiny burger, sitting on a park bench during autumn
\item a koala playing a ukulele in a eucalyptus tree, with soft evening light filtering through the leaves
\item a dog dressed as a superhero flying over a city, cape fluttering in the wind at dusk
\item a medieval castle perched on a misty mountain, torches flickering along its stone walls at night
\item a quaint cobblestone street lined with colorful houses, flower boxes blooming in a European village at sunrise
\item the Taj Mahal shimmering at dawn, its white marble reflecting in a still pool with lotus flowers
\item a cozy wooden cabin in a snowy forest, smoke curling from the chimney under a starry sky
\item an ancient Greek temple with marble columns, surrounded by olive trees under a golden sunset
\item a chef flipping pancakes in a bustling diner, steam rising as syrup drizzles over the stack
\item a close-up of a sushi chef slicing fresh salmon, rice rolling perfectly on a bamboo mat
\item an ice cream cone melting on a wooden table, colorful drips pooling in the summer heat
\item a Thanksgiving feast laid out on a grand table, turkey steaming beside golden rolls and cranberry sauce
\item a couple dancing romantically under a starry sky, their shadows swaying on a moonlit beach
\item a musician strumming a guitar by a campfire, friends singing along in a snowy forest
\item a photographer capturing the Northern Lights, camera clicking under a swirling green sky
\item a jogger running along a beach at sunrise, waves crashing gently beside them
\item a family picnicking in a park with autumn leaves, a red blanket spread with a basket of treats
\item a group of friends playing board games in a cozy cabin, rain tapping on the windows outside
\item a cyclist racing through a bustling city, tires humming on wet pavement after rain
\item a field of sunflowers swaying in a summer breeze, petals glowing under a clear blue sky
\item a giant redwood towering over a misty forest, sunlight streaming through its canopy
\item a canyon with red rock layers, a river winding through its depths at golden hour
\item a rainforest canopy teeming with colorful birds, mist clinging to the treetops
\item a tundra glowing under the Northern Lights, reindeer grazing on frosty ground
\item a serene lake reflecting autumn trees, a single canoe gliding across the mirror-like surface
\item a windswept prairie with tall grass, a lone tree standing against a dramatic sunset
\item a vintage car cruising down a country road, autumn leaves swirling in its wake
\item a hot air balloon drifting over a patchwork of fields, the sky a soft pastel at dawn
\item a helicopter hovering over a snowy mountain peak, rotors kicking up powdery snow
\item a steam locomotive puffing through a green valley, smoke curling into a clear sky
\item a futuristic hoverbike racing through a neon-lit city, reflections bouncing off wet streets
\item a boat sailing along the Seine River, the Eiffel Tower glowing in the background at dusk
\item a unicorn prancing through a rainbow-lit meadow, sparkles trailing in its wake
\item a talking owl perched on a glowing lantern, offering riddles in a foggy forest
\item a teddy bear rowing a tiny boat on a starry lake, fireflies dancing around it
\item a bustling harbor in the style of Impressionism, boats bobbing in soft, colorful strokes
\item a monochromatic forest in winter, snow blanketing trees in stark black and white
\end{enumerate}
}

\subsubsection{Temporal Dataset}
The Temporal subset foregrounds explicit motion and temporal progression, spanning locomotion, acceleration and deceleration, and articulated human and animal actions, thereby stressing a model's ability to synthesize coherent, physically plausible dynamics.

Prompt Template:
\begin{quote}
"As an expert linguist, your task is to create a high-quality textual dataset with \textbf{high dynamic content} for text-to-video generation tasks. Following the style and structure demonstrated in the reference examples provided below, generate 200 textual prompts. Ensure each prompt has low perplexity, clear semantics, and high comprehensibility, thus enabling \textbf{the generation of highly dynamic video content}. The prompts should cover diverse categories and scenarios, including but not limited to animals, architecture, food, humans, lifestyles, plants, scenery, and vehicles.

Reference samples:

a person swimming in ocean

a bicycle leaning against a tree

a car stuck in traffic during rush hour

\ldots
"
\end{quote}

Its 52 constituent prompts are:
{\small
\begin{enumerate}
\item a person washing the dishes
\item a bicycle accelerating to gain speed
\item a car stuck in traffic during rush hour
\item a car turning a corner
\item a car slowing down to stop
\item a motorcycle cruising along a coastal highway
\item a motorcycle accelerating to gain speed
\item a bus accelerating to gain speed
\item a train speeding down the tracks
\item a truck accelerating to gain speed
\item a dog playing in park
\item a dog running happily
\item a horse taking a peaceful walk
\item a horse running to join a herd of its kind
\item a sheep running to join a herd of its kind
\item a cow running to join a herd of its kind
\item an elephant running to join a herd of its kind
\item a bear catching a salmon in its powerful jaws
\item a zebra running to join a herd of its kind
\item a zebra taking a peaceful walk
\item a giraffe running to join a herd of its kind
\item a cargo ship navigating through stormy ocean waves
\item a kayak paddling swiftly through whitewater rapids
\item a horse-drawn carriage clip-clopping down a cobblestone street
\item a hang glider soaring above jagged cliff faces
\item a blacksmith hammering glowing metal in a fiery forge
\item a barista creating latte art in a cozy caf\'e
\item a florist arranging a bouquet of exotic tropical blooms
\item a skateboarder grinding along a metal handrail downtown
\item a bartender juggling cocktail shakers with flair
\item a kangaroo boxing with another kangaroo
\item a bartender shaking a cocktail mixer
\item a sushi chef rolling sushi
\item a barista grinding coffee beans
\item a person slicing vegetables for a salad
\item a runner sprinting in a race
\item a cyclist pedaling up a steep hill
\item a swimmer doing laps in a pool
\item a dancer performing ballet
\item a writer typing on a typewriter
\item a mechanic changing a tire
\item a hairdresser styling hair
\item a fisherman reeling in a catch
\item a group of friends singing karaoke
\item a person knitting a scarf
\item a teenager rollerblading in a skate park
\item a group of people doing a flash mob dance
\item a couple baking cookies together
\item a person walking barefoot on grass
\item a golf cart driving on a golf course
\item a segway touring a city
\item a rickshaw navigating through traffic
\end{enumerate}
}

\vspace{32pt}

\subsection{Additional Visualization Results}
\label{sup:vis}

\begin{figure*}[!ht]
    \vspace{2pt}
    \centering
    \includegraphics[width=0.98\linewidth]{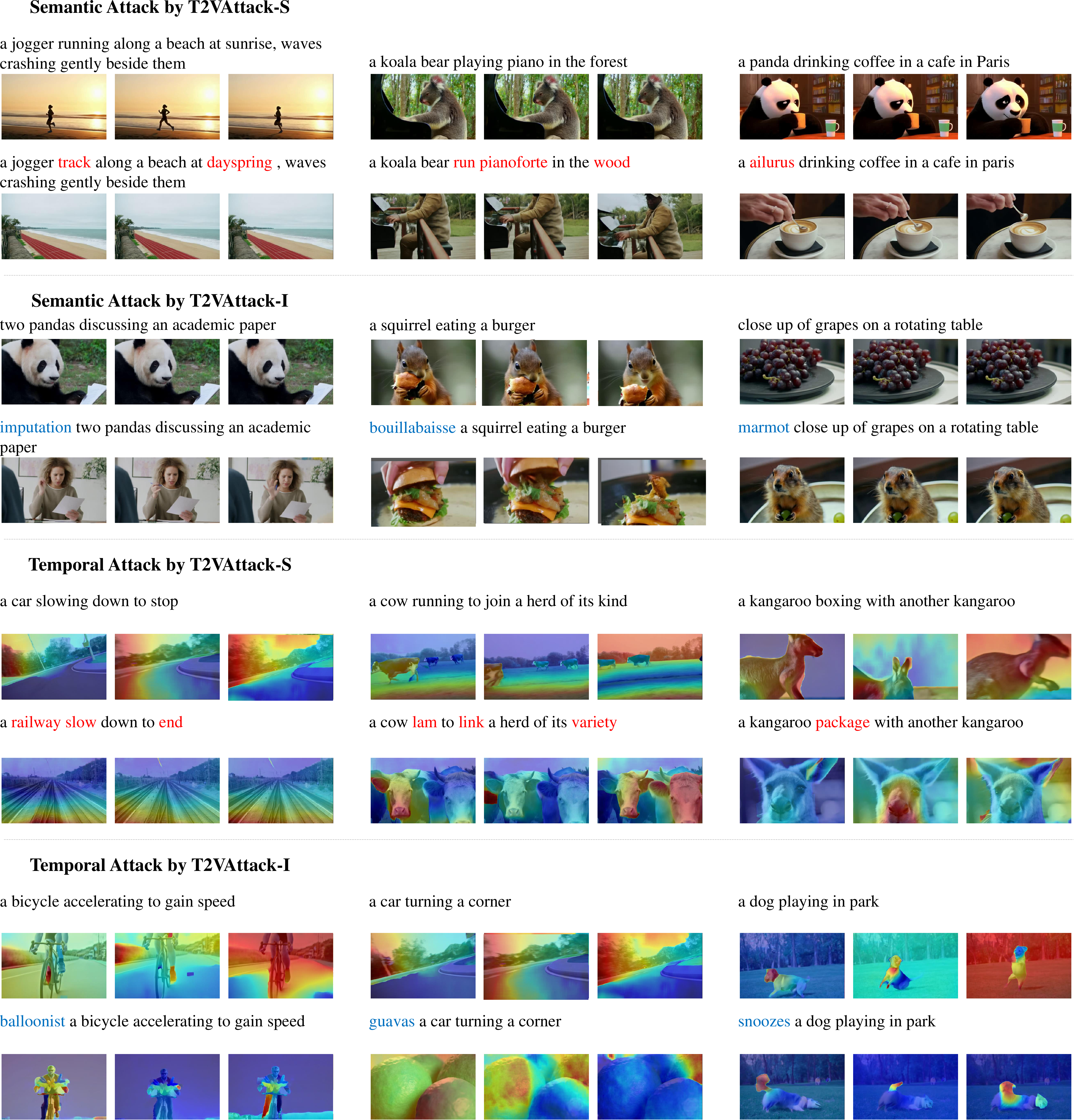}
    \caption{Visualization of T2VAttack-S and T2VAttack-I methods on HunyuanVideo. For each pair of videos, the top shows the video generated from the original prompt, while the bottom shows the video generated after applying an adversarial perturbation. The examples highlight that minor textual changes can induce substantial degradation in semantic fidelity and temporal dynamics. 
    }
    \label{fig:visualition-old}
\end{figure*}

Fig.~\ref{fig:visualition-old} provides additional visual examples illustrating the impact of T2VAttack-I and T2VAttack-S attacks on the \textbf{HunyuanVideo} model. Compared to other T2V models, HunyuanVideo demonstrates stronger robustness, making it more challenging to attack.
These results further validate the effectiveness of our proposed methods.

\end{document}